%% file: main.tex
\documentclass{article}

    \PassOptionsToPackage{numbers, compress}{natbib}

\usepackage{preprint}

\usepackage[utf8]{inputenc} 
\usepackage[T1]{fontenc}    
\usepackage{hyperref}       
\usepackage{url}            
\usepackage{booktabs}       
\usepackage{amsfonts}       
\usepackage{nicefrac}       
\usepackage{microtype}      
\usepackage{xcolor}         
\usepackage{float}
\usepackage{bbm}
\usepackage{todonotes}
\usepackage{multirow, makecell}
\usepackage{arydshln}
\usepackage{booktabs}
\usepackage{amsmath}
\usepackage{amssymb}
\DeclareMathOperator*{\argmax}{arg\,max}

\usepackage{listings}
\usepackage{xcolor}
\colorlet{punct}{red!60!black}
\definecolor{background}{HTML}{EEEEEE}
\definecolor{delim}{RGB}{20,105,176}
\colorlet{numb}{magenta!60!black}
\usepackage{multirow}
\usepackage{colortbl}
\usepackage{algorithm}
\usepackage{algpseudocode}
\usepackage[]{hyphenat}
\usepackage{textcomp}
\usepackage{microtype}
\usepackage{tcolorbox}
\usepackage{hyperref}
\usepackage{url}
\usepackage{multirow}
\usepackage{microtype}
\usepackage{hyperref}
\usepackage{url}
\usepackage{multirow}
\usepackage{wrapfig}

\usepackage{listings}
\usepackage[frozencache,cachedir=minted-cache]{minted}

\usepackage{xcolor,pifont}
\newcommand*\colourcheck[1]{%
  \expandafter\newcommand\csname #1check\endcsname{\textcolor{#1}{\ding{52}}}%
}
\colourcheck{ForestGreen}

\usepackage{caption}
\usepackage{subcaption}

\definecolor{light-gray}{gray}{0.9}

\newcommand{\dataname}[0]{\texttt{\textbf{DiGiT-TC}}}

\lstdefinelanguage{json}{
    basicstyle=\scriptsize\ttfamily,
    numbers=none,
    numberstyle=\scriptsize,
    stepnumber=1,
    numbersep=8pt,
    showstringspaces=false,
    breaklines=true,
    frame=lines,
    backgroundcolor=\color{background},
    literate=
     *{0}{{{\color{numb}0}}}{1}
      {1}{{{\color{numb}1}}}{1}
      {2}{{{\color{numb}2}}}{1}
      {3}{{{\color{numb}3}}}{1}
      {4}{{{\color{numb}4}}}{1}
      {5}{{{\color{numb}5}}}{1}
      {6}{{{\color{numb}6}}}{1}
      {7}{{{\color{numb}7}}}{1}
      {8}{{{\color{numb}8}}}{1}
      {9}{{{\color{numb}9}}}{1}
      {:}{{{\color{punct}{:}}}}{1}
      {,}{{{\color{punct}{,}}}}{1}
      {\{}{{{\color{delim}{\{}}}}{1}
      {\}}{{{\color{delim}{\}}}}}{1}
      {[}{{{\color{delim}{[}}}}{1}
      {]}{{{\color{delim}{]}}}}{1},
}

\title{Simulating Complex Multi-Turn Tool Calling \\ Interactions in Stateless Execution Environments}

\author{
  Maxwell Crouse, Ibrahim Abdelaziz, Kshitij Fadnis, Siva Sankalp Patel, Kinjal Basu, \\
  {\bf Chulaka Gunasekara, Sadhana Kumaravel, Asim Munawar, Pavan Kapanipathi} \\
  \texttt{\{maxwell.crouse, ibrahim.abdelaziz1\}@ibm.com} \\
  {\normalfont \textit{IBM Research AI}}
}

\begin{document}

\maketitle

\begin{abstract}
Synthetic data has proven itself to be a valuable resource for tuning smaller, cost-effective language models to handle the complexities of multi-turn tool calling conversations. While many frameworks and systems for producing synthetic multi-turn tool calling data have been proposed, prior works have frequently assumed that any tool calling interactions will take place in an execution environment that maintains state. When such an environment is available, this is advantageous as it allows for the validity of an interaction to be determined by whether or not the state of the execution environment matches to some pre-specified objective. Unfortunately, this does not hold in many real-world tool use settings, e.g., in enterprise settings where data security is of the utmost importance or in cases where tool specifications are synthesized from multiple sources. In this work, we address this gap by introducing a data generation method, \dataname\makeatletter
\@ifpackageloaded{preprint}{
\enlargethispage{2\baselineskip}
\footnote{Code for \dataname{} be found at \url{https://github.com/IBM/fms-dgt}}
}, that is designed to produce tool calling conversations that have the characteristics of conversations generated through search in a stateful environment. The key to our technique lies in a novel generation pattern that allows our approach to implicitly represent certain tool calls in the user request. We validate our approach on multiple standard multi-turn tool calling benchmarks and demonstrate that, even in stateful problem settings, our approach results in significant performance gains.
\end{abstract}

\input{sections/introduction}

\input{sections/related_work}

\input{sections/method}

\input{sections/experiments}
\input{sections/conclusion}

\bibliographystyle{plainnat}
\bibliography{references}

\clearpage
\appendix
\input{sections/appendix}
\makeatletter
\@ifpackageloaded{neurips_2026}{
\clearpage
\input{sections/checklist}
}
{}



\end{document}

%% file: sections/introduction.tex
\section{Introduction}
\label{sec:introduction}

Agentic applications of LLMs are in high demand due to their ability to effectively operate in a variety of environments that demand flexibility. While large, generalist models commonly serve as the backbone to such agents, their deployment can be fraught with challenges, including excessive costs \cite{belcak2025smalllanguagemodelsfuture}, privacy concerns \cite{iqbal2024llm}, and difficulties adapting to new domain-specific environments \cite{jimenez-gutierrez-etal-2022-thinking}. Smaller models tuned with synthetic data provide one such means of addressing these concerns, wherein the capabilities of a larger model are distilled into a smaller model through training on data custom to the domain of interest. However, synthetic data is not a panacea, and producing the right data for tuning smaller models without introducing new issues like catastrophic forgetting or overfitting remains an open problem \cite{shumailov2024ai,wenger2024ai}.

\begin{figure}[t]
\begin{tcolorbox}[
        colback=gray!5, 
        colframe=gray!70, 
        rounded corners,
        boxrule=1pt, 
        left=1mm, right=1mm, top=1mm, bottom=1mm
    ]
\begin{minipage}{0.4\textwidth}
\begin{tcolorbox}[colback=gray!30, colframe=gray!50, rounded corners, boxrule=0.5pt,left=1mm, right=1mm, top=1mm, bottom=1mm]
\begin{minted}[linenos=false,tabsize=4,fontsize=\scriptsize]{text}
{
  "role": "user",
  "content": "I want you to first check 
    for all movie showtimes for 'The 
    Avant-Garde Experiment' and then 
    book a ticket for the first 
    showtime you find"
}
\end{minted}
\end{tcolorbox}
\vskip 2pt
\begin{tcolorbox}[colback=gray!30, colframe=gray!50, rounded corners, boxrule=0.5pt,left=1mm, right=1mm, top=1mm, bottom=1mm]
\begin{minted}[linenos=false,tabsize=4,fontsize=\scriptsize]{text}
{
  "role": "user",
  "content": "I want to book a ticket 
    for the 'The Avant-Garde 
    Experiment'"
}
\end{minted}
\end{tcolorbox}
\end{minipage}%
\hfill
\begin{minipage}{0.59\textwidth}
\begin{tcolorbox}[colback=gray!10, colframe=gray!50, rounded corners, boxrule=0.5pt,left=1mm, right=1mm, top=1mm, bottom=1mm]
\begin{minted}[linenos=false,tabsize=4,fontsize=\scriptsize]{text}
{
  "role": "assistant",
  "content": {
    "name": "check_movie_showtimes",
    "arguments": { "title": "The Avant-Garde Experiment" },
    "id": "$1"
  }
}
{
  "role": "assistant",
  "content": {
    "name": "book_movie_tickets",
    "arguments": {
      "title": "The Avant-Garde Experiment",
      "show_time": "$1.showtimes[0]",
    },
    "id": "$2"
  }
}
\end{minted}
\end{tcolorbox}
\end{minipage}
\end{tcolorbox} 
\caption{Two examples of user requests (left), one where both tool calls are explicit (upper) and one that demonstrates an implicit tool call (lower), i.e., \texttt{check\_movie\_showtimes}}
\label{fig:implicit_tc}
\end{figure}

There are several different frameworks for producing synthetic data for tool calling \cite{liu2024apigen,yao2024tau,yin2025magnet,zeng2025toolace}. These frameworks will often operate over a suite of tools that are linked to some notion of a world state \cite{li-etal-2023-api, farn2023tooltalk,lu-etal-2025-toolsandbox}. That is, the LLM is interfacing with a service (e.g., a database) that has a set of stored data and can execute operations that update or access that data. 
In these frameworks, the objective is to take either a synthetic or manually curated user request that involves the world state and have the LLM execute actions via tool calls until it achieves the underlying goal of the request.

The path an LLM takes in solving a request, where the LLM is uncovering new information or transforming the underlying world state in order to achieve the user request, can result in complicated tool call trajectories in which many of the tool calls are \textit{implicit} (see Figure \ref{fig:implicit_tc}), i.e., not directly specified by the user's request \cite{basu2025nestfulbenchmarkevaluatingllms}. By keeping only those interactions where the final state matches the expected target, the risk of irrelevant intermediate tool calls and degenerate interactions is lessened.

Unfortunately, for many real-world tool calling settings where synthetic data could be of use, it is not always feasible to access a real backend with an accessible state. In enterprise domains where data security is a topmost concern, there are risks in giving LLMs privileged access to data or services \cite{Naihin2023TestingLM,ruanidentifying,Zhou2025SafeAgentSL}. Even in cases where a system's state is accessible, it might be expensive to interface with to produce data at scale, e.g., \cite{xie2024osworld}. Moreover, increasingly the top open language models incorporate synthetic, LLM-generated tools during training \cite{team2025kimi,zeng2025glm} which can, at best, only be simulated.

In this work we introduce \dataname{}, a framework for \textbf{D}ata \textbf{G}eneration and \textbf{T}ransformation for \textbf{T}ool-\textbf{C}alling, to address the problem of generating meaningfully complex, synthetic tool calling interactions from tool specifications alone and introduce a process for generating tool calls that imitate those generated through exploration within a stateful environment. The key insight of \dataname{} is to flip the generation of an interaction by prompting an LLM to first generate tool calls and then to generate a corresponding user request. This allows our method to be selective about which tool calls it should explicitly incorporate into the user request and which to leave implicit. As this step can introduce significantly noisy outputs, we reign in the quality of tool calling interactions through a back translation step that ensures the tool call sequence remains faithful to the user request.

To facilitate research in this area, we fully open-source all aspects of the system in this paper. Thus, both our data generation framework and data produced are available under a permissive Apache-2.0 license. Furthermore, all data generated by our approach is produced by \texttt{GPT-OSS} \cite{openai2025gptoss120bgptoss20bmodel}, a license-friendly, non-frontier, open-weights teacher model.

In summary, both the selective generation strategy and back translation for noise mitigation are novel contributions of \dataname{}. We evaluate the effectiveness of our approach on the multi-turn categories of the popular BFCL \cite{patilberkeley}, $\tau$-bench \cite{yao2024tau}, and ACEBench \cite{chen2025acebench} benchmarks where we demonstrate significant performance gains as compared to both base models and amongst comparably sized, fine-tuned models. Lastly, we fully open source all aspects of this work and set, to our knowledge, a new state-of-the-art performance across multi-turn benchmarks as compared to other models that also train using \textit{only} data produced by open-source, open-weights teacher models.

%% file: sections/related_work.tex
\section{Related Work}

\subsection{LLM Agents}
Building on tool-calling capabilities, recent research has focused on LLM-based agents that reason, plan, and act over multiple steps. Prompting strategies such as Chain-of-Thought \cite{wei2022chain}, ReAct \cite{yao2023react}, and Reflection \cite{shinn2023reflexion} enable iterative decision-making, which has inspired frameworks like LangChain\footnote{https://www.langchain.com/}, AutoGPT\footnote{https://agpt.co/}, and BabyAGI\footnote{https://babyagi.org/} to orchestrate multi-turn tool use for dynamic tasks such as information retrieval, web navigation, and API-based service composition \cite{nakano2021webgpt, shinn2023reflexion, zhang2023data}. 

Popular benchmarks for evaluating agents and tool calling models include BFCL \cite{patilberkeley}, StableToolBench \cite{guo2024stabletoolbench}, $\tau$-Bench \cite{yao2024tau}, and ACEBench \cite{chen2025acebench} have been introduced to evaluate such agents and tool-augmented LLMs across single and multi-turn scenarios. Both BFCL and ACEBench evaluate LLMs on a breadth of multi-turn, tool-augmented reasoning tasks while $\tau$-bench tests in a dual-control settings where both the agent and a simulated user act on a shared environment.

\subsection{Synthetic Data for Tool-Calling}

High-quality training data has been a critical driver of LLM performance, yet the scarcity of such datasets for tool-using agents has motivated research on synthetic data generation. Early efforts primarily focused on single-turn synthesis, where a model receives a one-time query and responds accordingly \cite{patilberkeley,liu2024apigen,qin2023toolllm}. However, real-world tasks often involve multi-turn interactions, prompting recent studies to explore conversational data synthesis. 

In the multi-turn setting, BUTTON \cite{chen2024facilitating} generates compositional instruction data via multi-agent setups, though it lacks systematic quality control. Approaches such as MAGNET \cite{yin2025magnet} and ToolDial \cite{shim2025tooldial} leverage structured, graph-based representations to generate iterative query-call sequences and multi-turn dialogues, capturing rich user-system interactions and dynamic API usage. ToolACE-MT \cite{zeng2025toolace} introduces a generation framework that simulates an entire conversational trajectory in one shot and then iteratively refines it to inject known patterns of complexity. Similarly, APIGen-MT \cite{prabhakar2025apigen} uses a two-phase process in which task blueprints with ground-truth actions are first generated and reviewed by LLMs, then transformed into full multi-turn interaction trajectories via simulated human-agent interplay. Lastly, TOUCAN \cite{xu2025toucan} is a recent approach that synthesizes tool calling data from agentic trajectories that involve near 500 real-world Model Context Protocols (MCPs). Unlike other approaches, TOUCAN places a distinct emphasis on breadth of coverage and leverages real MCP environments to produce their data. 

The aforementioned works share many goals and high-level details in common with our approach (e.g., ToolACE-MT \cite{liutoolace} also does not assume access to an execution environment). However, our work differs in its focus on deriving a repeatable process for producing complex conversations with many implicit tool calls in the absence of an execution backend with a meaningful state, i.e., conversations where correct tool call sequences do not directly correspond to user requests.

\begin{figure}[t]
\centering
\includegraphics[width=0.9\columnwidth]{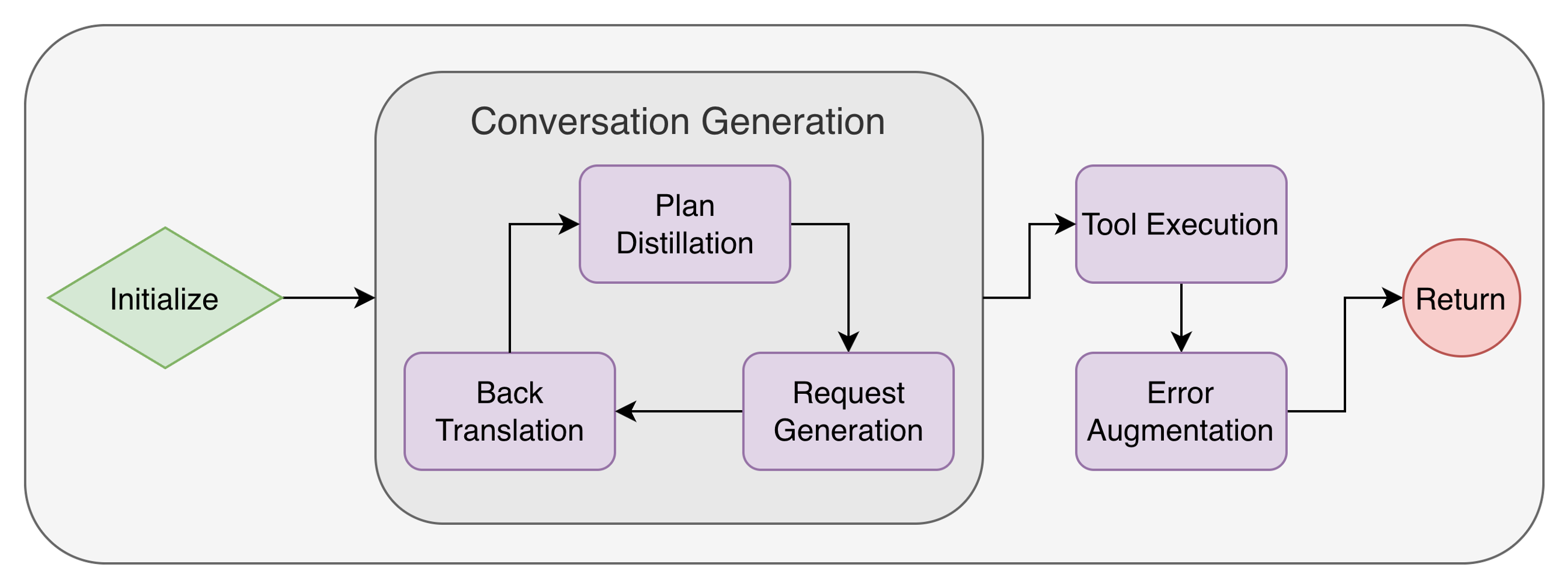} 
\caption{Conversation generation stages in \dataname{}}
\label{fig:mt_sdg_overview}
\end{figure}

%% file: sections/method.tex
\begin{figure*}[t]
\centering
\includegraphics[width=0.8\textwidth]{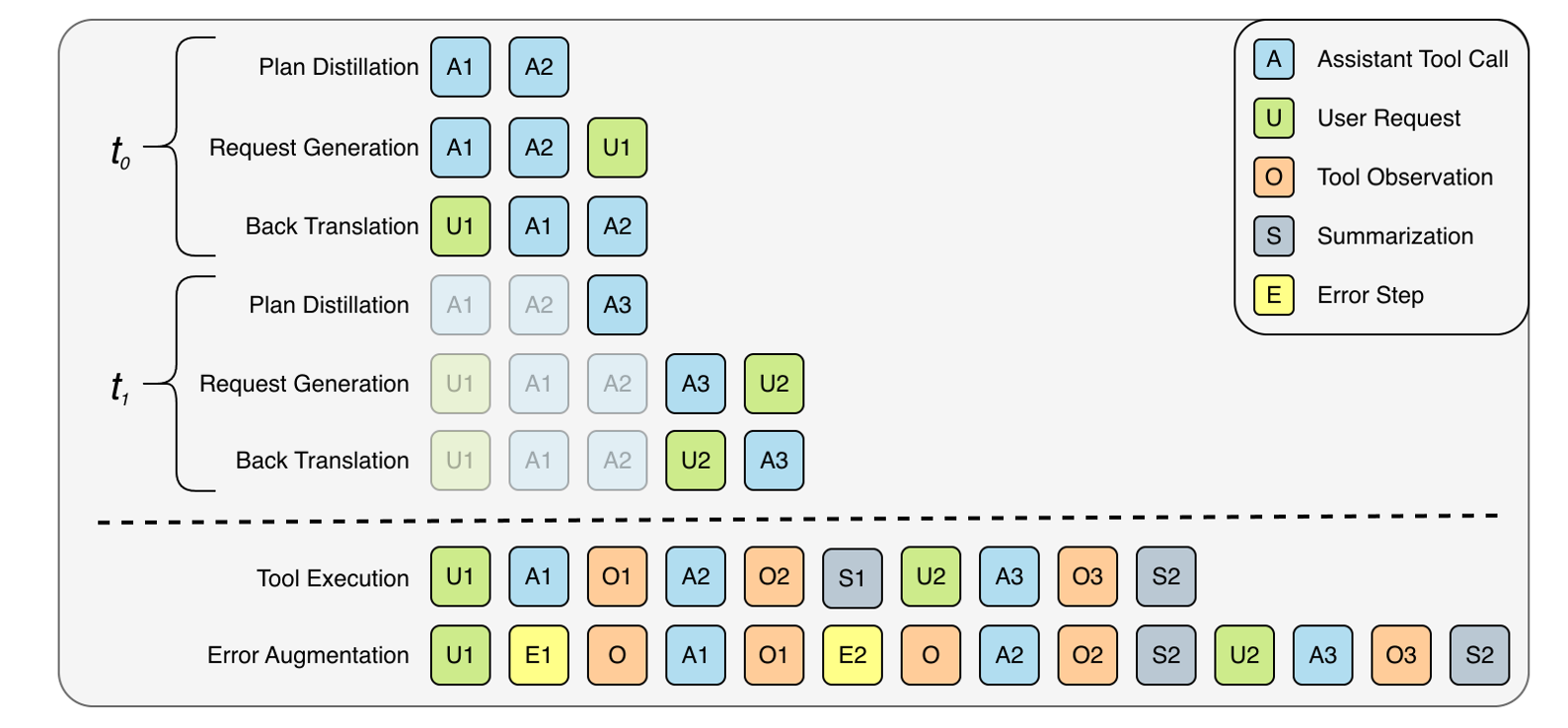} 
\caption{Construction of a two-turn conversation}
\label{fig:mt_sdg_turns}
\end{figure*}

\section{Method}

The objective of \dataname{} is to construct high quality, multi-turn interactions that consist of a "user" making requests to an "assistant" that must be solved through the use of tools. To accomplish this, our approach assumes only that it is provided a set of tool specifications, where each tool specification contains a description of the tool itself and its input and output parameters (an example specification is provided in the Appendix).

\dataname{} constructs a tool calling conversation in stages. The core of data generation occurs in a back translation loop, which iteratively adds "user" and "assistant" turns until a prespecified number of iterations have completed. To aid in describing our approach, we provide Figure \ref{fig:mt_sdg_overview} of a simple diagram illustrating the flow of our framework and Figure \ref{fig:mt_sdg_turns} which depicts how turns are added by each stage to the final output. In the following sections, we will go through each stage in detail and explain how they contribute to the overall effectiveness of our approach. All prompts used in each stage and a complete output are provided in the Appendix.


\subsection{Initialization}

Each conversation is initialized as a tuple $\left< M, T, \mathcal{U}\right>$ consisting of an empty list of messages $M$, a set of tools $T$ sampled from all available tools, and a user persona $\mathcal{U}$. From the persona and sampled set of tools, our approach augments $\mathcal{U}$ with more details of the user by prompting an LLM to generate specific details of the fictitious user that are relevant to the sampled tools in $T$ (e.g., their location, profession, etc.). Once the LLM has completed generation, this tuple of information is passed into the main loop of \dataname{}, which will generate the turns of the interaction.

\subsection{Conversation Generation}
\label{sec:conv_gen}

The core of \dataname{} is the iterative loop in which "user" and "assistant" turns are generated and added to the messages $M$. On each iteration, our approach will first produce a tool call sequence based on the user details and tools available and then generate a corresponding user utterance. The tool call sequence predicted will remain unexecuted, i.e., the tool calls will contain inter-tool references that have yet to be filled in.

By flipping how tool calling interactions are constructed, our approach can have finer-grained control over what parts of the tool call sequence should be explicitly represented in the user request. Once generated, our method ensures quality by performing verification of the tool call sequence with a back translation step. In the next sections, we will walk through the three stages of this phase of generation in the order they are executed by \dataname{}.

\begin{figure}[t]

\begin{tcolorbox}[
        colback=gray!5, 
        colframe=gray!70, 
        rounded corners,
        boxrule=1pt, 
        left=1mm, right=1mm, top=1mm, bottom=1mm
    ]
\begin{minipage}{0.498\textwidth}
\begin{tcolorbox}[colback=gray!10, colframe=gray!50, rounded corners, boxrule=0.5pt,left=1mm, right=1mm, top=1mm, bottom=1mm]
\begin{minted}[linenos=false,tabsize=4,fontsize=\scriptsize]{text}
{
  "name": "get_movie_genre",
  "arguments": {
    "title": "The Avant-Garde Experiment"
  },
  "id": "$1"
},
{
  "name": "check_movie_showtimes",
  "arguments": {
    "title": "The Avant-Garde Experiment"
  },
  "id": "$2"
},
{
  "name": "book_movie_tickets",
  "arguments": {
    "title": "The Avant-Garde Experiment",
    "show_time": "$2.showtimes[0]",
  },
  "id": "$3"
}
\end{minted}
\end{tcolorbox}
\end{minipage}%
\hfill
\begin{minipage}{0.498\textwidth}
\begin{tcolorbox}[colback=gray!10, colframe=gray!50, rounded corners, boxrule=0.5pt,left=1mm, right=1mm, top=1mm, bottom=1mm]
\begin{minted}[linenos=false,tabsize=4,fontsize=\scriptsize]{text}
{
  "name": "check_movie_showtimes",
  "arguments": {
    "title": "The Avant-Garde Experiment"
  },
  "id": "$1"
},
{
  "name": "book_movie_tickets",
  "arguments": {
    "title": "The Avant-Garde Experiment",
    "show_time": "$1.showtimes[0]",
  },
  "id": "$2"
}

\end{minted}
\end{tcolorbox}
\end{minipage}
\end{tcolorbox}
\caption{Example of distillation where singleton elements (left query) are filtered out (right query)}
\label{fig:plan_distill}
\end{figure}

\subsubsection{Plan Distillation}

Following initialization, tool call generation begins from the tuple $\left< M, T, \mathcal{U}\right>$. Our framework instructs an LLM to produce each tool call sequence $S = \langle s_1, \ldots, s_n \rangle$ from the tools $T$ with arguments based on the user details $\mathcal{U}$. In addition, the LLM is instructed to write tool calls within $S$ that reference one another through Pythonic-style references \cite{basu2025nestfulbenchmarkevaluatingllms}, with the inner contents of tool call outputs being accessible through dot notation and list accessors.

Because we have generated a tool call sequence in isolation from a user utterance, each sequence $S$ may contain many tool calls that do not serve a clear purpose. As such, the next step in preparing the final tool call sequence is to distill from $S$ a more concise subsequence of tool calls with certain properties we would desire in our conversation.

Our approach treats the sequence $S$ as a directed-acyclic graph (DAG), where edges within the DAG are defined by inter-tool references. From the DAG, it pulls the set of all connected components $C$ and selects the largest component to serve as the final tool call for this iteration, i.e.,
\begin{equation*}
    S = \argmax_{C_i \in C} |C_i|
\end{equation*}
Figure \ref{fig:plan_distill} provides an example of this distillation process, where an individual tool call sequence is distilled into a more compact form.

\subsubsection{Request Generation}
\label{sec:req_gen}

In this stage, the tool call sequence $S$ is translated into a corresponding user utterance. Provided with both our tuple $\left< M, T, \mathcal{U} \right>$ from initialization and the tool call sequence $S$ from the prior stage, an LLM is prompted to produce a user request $r$ that would be solved by executing the tool calls in $S$. 

\begin{wrapfigure}[14]{L}{0.55\textwidth} 
  \begin{minipage}{0.55\textwidth}
    \begin{algorithm}[H]
    \caption{Implicit Tool Call Selection}
    \label{alg:impl_sel}
    \begin{algorithmic}[1] 
        \State Initialize $I \leftarrow \emptyset$
        \State $S \leftarrow \{\ s \in S : \exists p \in S . \ s \in \texttt{children}(p, S)\ \}$
        \State $r \in_R \{1, \ldots, |S|\}$
        \While{$|I| < r$}
            \State $c \in_R \{s \in S : \texttt{children}(s, S) = \emptyset\}$
            \State $I \leftarrow I \cup \{c\}$
            \State $S \leftarrow S \setminus \{c\}$
        \EndWhile
        \State \Return $I$
    \end{algorithmic}
\end{algorithm}
\end{minipage}
\end{wrapfigure}
As highlighted in Section \ref{sec:introduction}, in order to generate complex tool calling interactions, our approach seeks to produce tool call sequences with implicit tool calls. To achieve this, our method first partitions $S$ into two separate sequences $I$ and $E$, for implicit and explicit tool call sequences, respectively, with the partitioning done randomly to ensure the sizes of both $I$ and $E$ vary across examples.

Algorithm \ref{alg:impl_sel} provides our straightforward solution to this problem. The algorithm picks a target size for the set $I$, removes any element of $S$ that is not a dependency to some other element, and then iteratively fills the set with elements in $S$ chosen at random, where the chosen elements are constrained to \textit{only} have dependencies (i.e., children) from other elements of $I$. Upon completion the algorithm returns $I$, and the complement of $I$ in $S$ is used as $E$, i.e., $E = S \setminus I$. The LLM is then prompted to treat the tool call sequence $E$ as the explicit goal of the user's request and to avoid directly mentioning any of the tool calls in $I$ (the exact prompts are provided in the Appendix).

Refering back to our example in Figures \ref{fig:implicit_tc} and \ref{fig:plan_distill}, the output of the first tool call, \texttt{\$1.showtimes}, is used as an argument to the second tool call. In this case, $I$ would be constructed as the singleton containing \texttt{check\_movie\_showtimes} tool call, which would then result in $E$ containing the \texttt{book\_movie\_tickets} tool call.

Once a user request has been generated, our method validates the tool call sequence is not a straightforward transcription of the user request using the Kendall's $\tau_b$ correlation coefficient. First, we define the rank of a particular argument node $l_i$ to be the indicator variable $R_{tc}[l_i]$ for whether or not its parent tool call appears in the hidden set. We further write the rank of a leaf node as it occurs in the user request to be $R_{req}[l_i]$ (i.e., the order it appears in the request as compared to other leaves). In this context, each leaf node $l_i$ corresponds to a single observation $(x_i, y_i) = (R_{tc}[l_i], R_{req}[l_i])$. We write a pair of observations as $(x_i, y_i)$ and $(x_j, y_j)$ and consider them concordant when
\begin{equation*}
    \left( x_i < x_j \wedge y_i < y_j \right) \vee \left( x_i > x_j \wedge y_i > y_j \right)
\end{equation*}
Similarly, the pair is said to be discordant when
\begin{equation*}
    \left( x_i < x_j \wedge y_i > y_j \right) \vee \left( x_i > x_j \wedge y_i < y_j \right)
\end{equation*}
From these two notions, we calculate the Kendall's $\tau_b$ coefficient as
\begin{equation*}
    \tau_b = \dfrac{C - D}{\sqrt{(C + D + X_0)(C + D + Y_0)}}
\end{equation*}
where $C$ is the number of concordant pairs, $D$ the number of discordant pairs, $X_0$ the number of pairs where the first element is tied, and $Y_0$ the number of pairs where the second element is tied.

When $\tau_b$ indicates the orderings of arguments between the tool calls and the user request are correlated (i.e., $\epsilon > 0$), our approach filters out the example. Intuitively, this gives stronger preference to examples where the tool call sequence is not a straightforward linear transcription of the user request.

\subsubsection{Back Translation}
\label{sec:back_trans}

To increase the quality of outputs, \dataname{} performs a back translation step that validates whether or not a new tool call sequence, $S^\prime$, predicted from the user request, $r$, and tools, $T$, would independently produce similar results to the original tool call sequence $S$. 

We observed it to be helpful to have a tolerance for mismatches, and thus validating consistency between the original tool call sequence, $S$, and new tool call sequences, $S^\prime$, involves only checking that the leaves of $S$ are all represented in $S^\prime$, i.e.,
\begin{align*}
    &\texttt{leaves}(S)\setminus \texttt{leaves}(S^\prime) = \emptyset
\end{align*}
When this condition fails, this is treated as an indication of noise in the generation process and the example is filtered out. Assuming $S^\prime$ has passed validation, it is added to the list of messages $M$ as an assistant step in lieu of $S$.

\begin{figure}[t]
\begin{tcolorbox}[
        colback=gray!5, 
        colframe=gray!70, 
        rounded corners,
        boxrule=1pt, 
        left=1mm, right=1mm, top=1mm, bottom=1mm
    ]
\begin{minipage}{0.498\textwidth}
\begin{tcolorbox}[colback=gray!10, colframe=gray!50, rounded corners, boxrule=0.5pt,left=1mm, right=1mm, top=1mm, bottom=1mm]
\begin{minted}[linenos=false,tabsize=4,fontsize=\scriptsize]{text}
{
  "role": "user",
  "content": "What's the monthly for a $35,000 
    car with a 5.2% annual interest rate."
}
{
  "role": "assistant",
  "content": "Could you please specify the down 
    payment amount and the loan term in months?"
}
{
  "role": "user",
  "content": "Sure, that's $8,000 and 60 months."
}
{
  "role": "assistant",
  "content": {
    "name": "compute_vehicle_payments",
    "arguments": {
      "vehicle_cost": 35000,
      "initial_deposit": 8000,
      "duration_months": 60,
      "annual_rate_percent": 5.2
  }
}
\end{minted}
\end{tcolorbox}
\end{minipage}%
\hfill
\begin{minipage}{0.498\textwidth}
\begin{tcolorbox}[colback=gray!10, colframe=gray!50, rounded corners, boxrule=0.5pt,left=1mm, right=1mm, top=1mm, bottom=1mm]
\begin{minted}[linenos=false,tabsize=4,fontsize=\scriptsize]{text}
{
  "role": "assistant",
  "content": {
    "name": "log_calorie_intake",
    "arguments": {
      "energy_intake": 0
    }
  }
}
{
  "role": "tool",
  "content": "The calorie count should 
    reflect actual intake rather than zero"
}
{
  "role": "assistant",
  "content": {
    "name": "log_calorie_intake",
    "arguments": {
      "energy_intake": 2120
    }
  }
}

\end{minted}
\end{tcolorbox}
\end{minipage}
\end{tcolorbox}
\caption{Example of a clarification error (left) and a tool call error (right)}
\label{fig:err_aug_tc}
\end{figure}

\subsection{Termination}

Once conversation generation is complete, the list of messages $M$ now consists of an alternating sequence of user requests and tool calls. Before returning to the user, \dataname{} finishes with two stages (see the last two stages of Figure \ref{fig:mt_sdg_turns}). The first of these stages involves executing the tool calls and substituting the values for each variable into their respective arguments and adding an "assistant" summary of the tool call results as they relate to the original request. The last step before returning, as with \cite{zeng2025toolace}, is to add complexity by simulating errors.

\subsubsection{Tool Execution and Summarization}

The tool call sequences must be executed, as, thus far, the tool calls contain references that link their outputs to the inputs of other tool calls. Our method simulates the execution of each tool call with an LLM to obtain a tool observation $O$. This is used to substitute required results into the appropriate slots of other tool calls in $M$ that have a dependency on $S$. After adding the result of tool call execution after $S$ in the list of messages $M$, the LLM is prompted to summarize the results of that interaction before moving on to the next user utterance.

\subsubsection{Error Augmentation}
\label{sec:error_handling}
\dataname{} constructs data to represent both human errors as well as tool execution errors. Human errors are considered those where the user asks an unsolvable or incorrect request, and the assistant must ask for clarification to determine the correct request. These are simulated by prompting an LLM to generate overly generic, unsolvable requests that either involve missing required parameters or insufficient tools in the interaction and then to generate the corresponding clarification question. In Figure \ref{fig:err_aug_tc} we provide an example of the interaction produced for this type of error.

Formulating a meaningful tool execution error is more challenging because there is no world state underlying the execution of tools in our approach. Here, we make the straightforward assumption that the purpose of an error message is to direct the model towards the correct solution. To simulate an error, our method takes in the list of messages $M$ that ends with a target tool call $S$ and then predicts a tool call $E$. When $S$ and $E$ differ, our approach prompts an LLM to suggest a hint for correcting the difference between the gold tool call $S$ and the predicted tool call $E$. This hint becomes the error message for a tool call observation $O$ that directly follows $E$ in $M$ (see Figure \ref{fig:err_aug_tc} for an example).

%% file: sections/experiments.tex
\begin{table*}[ht!]
\setlength{\tabcolsep}{3pt} 
\footnotesize
\centering
\begin{tabular}{@{}lccccccc@{}}
\toprule
\multirow{4}{*}{\textbf{\small System}} & \multirow{4}{*}{\textbf{\small Base Model}} & \multirow{4}{*}{\textbf{\small Teacher Model}} & \multicolumn{5}{c}{\textbf{Multi-Turn Category}} \\
\cmidrule(lr){4-8}
 & & & 
\multicolumn{1}{c}{\textit{Overall}} &
  \multicolumn{1}{c}{\textit{Base}} &
  \multicolumn{1}{c}{\begin{tabular}[c]{@{}c@{}}\textit{Miss}\\ \textit{Func}\end{tabular}} &
  \multicolumn{1}{c}{\begin{tabular}[c]{@{}c@{}}\textit{Miss}\\ \textit{Param}\end{tabular}} &
  \multicolumn{1}{c}{\begin{tabular}[c]{@{}c@{}}\textit{Long}\\ \textit{Context}\end{tabular}} \\ 
  \midrule
\textbf{GPT-4o-2024-11-20} \cite{hurst2024gpt} & -- & -- & 50.0 & 61.0 & 45.5 & 35.5 & 58.0 \\
\textbf{Gemini-2.5-Pro-05-06} \cite{reid2024gemini} & -- & -- & 34.6 & 39.5 & 29.5 & 31.5 & 38.0 \\
\midrule
\textbf{DeepSeek-V3-0324} \cite{liu2024deepseek} & -- & -- & 29.9 & 41.0 & 21.0 & 23.0 & 34.5 \\
\textbf{Llama3.1-70B-Inst} \cite{llama-3} & -- & -- & 12.5 & 17.0 & 13.0 & 10.5 & 9.5 \\
\textbf{Qwen3-8B} \cite{yang2025qwen3} & -- & -- & 37.0 & 42.5 & 38.5 & 31.5 & 35.5 \\
\textbf{Llama3.1-8B-Inst} \cite{llama-3} & -- & -- & 9.3 & 12.0 & 10.0 & 7.0 & 8.0 \\
\textbf{Qwen2.5-7B-Inst} \cite{Yang2024Qwen25TR} & -- & -- & 11.5 & 13.5 & 14.5 & 11.0 & 7.0 \\
\textbf{Granite-4.1-3B} \cite{granite2026} & -- & -- & 22.5 & 28.5 & 19.5 & 19.0 & 23.0 \\
\midrule
\textbf{ToolACE-MT-8B} \cite{liutoolace} & \textit{Llama3.1-8B-Inst} & \texttt{GPT-4o} & 40.3 & 57.5 & 31.5 & 34.0 & 38.0 \\
\textbf{MAS} \cite{liutoolace} & \textit{Llama3.1-8B-Inst} & \texttt{GPT-4o} & 31.4 & 46.5 & 19.0 & 31.0 & 29.0 \\ 
\textbf{Magnet} \cite{yin2025magnet} & \textit{Qwen-Coder-7B-Inst} & \texttt{Gemini-1.5-Pro} & 26.5 & 35.5 & 24.0 & 27.5 & 19.0 \\
\textbf{ToolWeave} \cite{khandelwal2026toolweave} & \textit{Llama3.1-8B-Inst} & \texttt{GPT-OSS} & 25.0 & 31.5 & 20.5 & 24.5 & 23.5 \\
\textbf{Toucan-1.5M} \cite{xu2025toucan} & \textit{Qwen2.5-7B-Inst} & \texttt{GPT-OSS/Kimi-K2/Qwen3} & 22.6 & NR & NR & NR & NR \\
\midrule
\multirow{3}{*}{\dataname{} (ours)} & \textit{Granite-4.1-3B} & \multirow{3}*{ \texttt{GPT-OSS}} & 26.6 & 39.5 & 18.0 & 16.5 & 32.5 \\
 & \textit{Qwen2.5-7B-Inst} &  & 24.5 & 37.0 & 19.5 & 24.0 & 17.5 \\
 & \textit{Llama3.1-8B-Inst} &  & 35.5 & 43.5 & 27.5 & 30.5 & 40.5 \\
\bottomrule
\end{tabular}
\caption{Multi-Turn BFCL accuracy results (\%) as compared to (1) frontier models, (2) open models, and (3) prior work trained from frontier / open models data (NR -- Not Reported)}
\label{res:bfcl_overall}
\vskip 10pt
\begin{tabular}{@{}lccccc@{}}
\toprule
\multirow{3}{*}{\textbf{\small System}} & \multirow{3}{*}{\textbf{\small Base Model}} & \textbf{BFCL} & \textbf{ACEBench} & \multicolumn{2}{c}{\textbf{$\tau$-bench}} \\
\cmidrule(lr){3-3}
\cmidrule(lr){4-4}
\cmidrule(lr){5-6}
 & & \textit{Multi-Turn} & \textit{Multi-Turn} &
\multicolumn{1}{c}{\textit{Retail}} &
\multicolumn{1}{c}{\textit{Airline}} \\
\midrule
\textbf{Llama3.1-8B-Inst} & \multirow{6}{*}{\textit{\small Llama3.1-8B-Inst}}& 9.3 & 24.0 & 6.1 & 26.0 \\
\textbf{Toucan-1.5M (sampled 8k)}$^*$ & & 15.0 & 41.0 & 18.3 & 18.0 \\
\textbf{ToolACE-MT-8B} & & 40.3 & 51.0 & 25.2 & 16.0 \\
\textbf{MAS} & & 31.4 & 48.0 & 21.7 & 10.0 \\
\textbf{ToolDial-9k} & & 12.3 & 30.0 & 10.4 & 20.0 \\
\textbf{\dataname{}} (ours) & & 35.5 & 46.0 & 14.7 & 23.0 \\
\bottomrule
\end{tabular}
\caption{Accuracy comparison (\%) on multi-turn categories of standard benchmarks \cite{patilberkeley,liutoolace,yao2024tau}. $^*$Results for \texttt{Toucan-1.5M} are those reported in \cite{liutoolace}}
\label{res:tauacebench_overall}
\end{table*}

\section{Experimental Evaluation}

To generate training data, we produced multi-turn instances using the license-friendly, open source \texttt{GPT-OSS-120B} \cite{openai2025gptoss120bgptoss20bmodel} model with source tools taken from \cite{yang2025toolmindtechnicalreportlargescale}. Like \cite{zeng2025toolace,prabhakar2025apigen}, we applied a basic LLM-as-a-judge filter to remove noisy conversations that may have made it through generation (the prompt is provided in the appendix). After filtering, training conversations concatenated together at random (to a maximum of 10 conversations per instance) to produce longer conversations, resulting in a final training dataset with 5,000 data points. To add variety, function names were masked with random strings to ensure the model did not overfit to specific values. We provide further detail on the training hyperparameters, evaluation setup, and more data analyses in the appendix.

For our main results, we use the multi-turn category of the popular BFCL benchmark to determine (1) the magnitude of improvement over the base model and (2) assess the impact of each of \dataname{}'s primary contributions. For reference, we also include results from popular base models and specialized fine-tuned models, provided that they are self-contained and do not pull data from external works (e.g., APIGen-MT \cite{prabhakar2025apigen} which achieves SOTA for BFCL but incorporates an unreleased blend of agentic data from \cite{zhang-etal-2025-xlam,zhang-etal-2025-actionstudio,liu2024apigen}). To validate the generalizability of our approach, we also evaluate the trained model on $\tau$-bench \cite{yao2024tau} and the multi-turn category of ACEBench \cite{liutoolace}.

\section{Results and Discussion}

We provide our main results for BFCL in Table \ref{res:bfcl_overall}. In the table, it can be seen that our approach performs strongly, exceeding the performance of DeepSeek \cite{liu2024deepseek} and the frontier Gemini model \cite{reid2024gemini}. Importantly, the data generated by our approach significantly boosted (ranging from 6 to 25 points) the performance of the underlying base model, regardless of the base model chosen. As compared to prior fine-tuned tool calling approaches, ours outperforms the recent works of Toucan \cite{xu2025toucan}, Magnet \cite{yin2025magnet}, MAS \cite{zeng2025toolace}, and ToolWeave \cite{khandelwal2026toolweave} for the results reported with similar sized models. Our approach produces the best results by far among those methods that derive data exclusively from open models.

To demonstrate generalizability, we provide Table \ref{res:tauacebench_overall} which shows our results across the multi-turn categories of three benchmarks. Overall, \dataname{} again demonstrates strong performance across benchmarks, with better performance than prior approaches built on open data. These results highlight the robustness of our method and the broader importance of data containing implicit tool calls.

 Though difficult to determine exactly the cause of the performance difference between our approach and that of ToolACE-MT \cite{zeng2025toolace}, we suspect our use of \texttt{GPT-OSS-120B} \cite{openai2025gptoss120bgptoss20bmodel} instead of the much more powerful \texttt{GPT-4o} \cite{hurst2024gpt} had at least some effect. As shown in both Magnet \cite{yin2025magnet} and ToolACE-MT \cite{zeng2025toolace}, using more powerful teacher models can result in drastic performance improvements. Unfortunately, the ability to use these powerful frontier models is often limited by their cost \cite{chenfrugalgpt}, privacy and data-security handling \cite{chen2025survey,desai2024opportunities}, and also the  legal restrictions on the use of their outputs to build models and systems (see the OpenAI\footnote{https://openai.com/policies/terms-of-use/} and Anthropic\footnote{https://www.anthropic.com/legal/consumer-terms} terms of use).

\begin{table*}[]
\footnotesize
\centering
\begin{tabular}{@{}lccccc@{}}
\toprule
\multirow{3}{*}{\textbf{\small System}} & \multicolumn{5}{c}{\textbf{Multi-Turn Category}} \\
\cmidrule(lr){2-6}
 & 
\multicolumn{1}{c}{\textit{Overall}} &
  \multicolumn{1}{c}{\textit{Base}} &
  \multicolumn{1}{c}{\textit{Miss Func}} &
  \multicolumn{1}{c}{\textit{Miss Param}} &
  \multicolumn{1}{c}{\textit{Long Context}} \\
  \midrule
\bf{\dataname{}} ($2k$) & 28.5 & 41.0 & 20.0 & 22.0 & 31.0 \\      
\bf{\ \ \ \ -- BT} & 25.4 & 39.0 & 19.0 & 19.5 & 24.0 \\
\bf{\ \ \ \ -- Implicit} & 24.5 & 37.0 & 20.0 & 21.0 & 20.0 \\
\bf{\ \ \ \ -- Error} & 23.8 & 35.0 & 19.0 & 19.0 & 22.0 \\
\bf{\ \ \ \ + \texttt{GPT-OSS-20B}} & 26.4 & 39.5 & 19.0 & 21.0 & 26.0 \\
\bottomrule
\end{tabular}
\caption{Ablation results for the Multi-Turn category of BFCL. In all experiments, $2k$ examples were used for training}
\label{res:bfcl_v3_abl_subtr}
\end{table*}

\subsection{Ablation Experiments}

We performed ablation experiments using the multi-turn categories of the BFCL dataset to validate the effectiveness of the various components of this work. To test the value of implicit tool calls, we trained a model on data where no elements of any tool call sequence were masked (see Section \ref{sec:req_gen}), thus making each user request a straightforward translation of its corresponding tool call sequence. Similarly, for back translation and error augmentation, we trained models on data that skipped the back translation and error augmentation steps, respectively (see Sections \ref{sec:back_trans} and \ref{sec:error_handling}). Lastly, to assess the impact of teacher model size, we also trained a model on data produced and filtered by \texttt{GPT-OSS-20B}. Importantly, in \textit{all} settings, the $2k$ examples were filtered using the same LLM-as-a-judge strategy for our main experiments.

Table \ref{res:bfcl_v3_abl_subtr} shows the results of these experiments, where it can be seen that both contributions had significant impact on the final results. Overall, excluding implicit tool calls led to a steep drop of 4\% in performance, which supports our hypothesis that implicit tool calls are an important feature to capture. Excluding back translation resulted in a similar drop of 3\%. Interestingly, this drop occurs even though the data has been filtered by an LLM for quality, which we suspect is due to the inherent noise of the initial planning step which produces outputs ungrounded by any user request.

Using a smaller teacher model resulted in drop of around 2\%, a lower but still reasonably high level of performance. This is not entirely surprising, as \texttt{GPT-OSS-20B} has been shown to often match (or even exceed) \texttt{GPT-OSS-120B} across several benchmarks \cite{bi2025gpt}. For lower-resource practitioners, we note that \texttt{GPT-OSS-20B} can be freely downloaded and run on a consumer laptop. Thus, renting dedicated GPUs or interfacing with costly LLM service providers is not required for those wishing to run our system locally.

\section{Limitations}

While we believe \dataname{} provides strong advantages from a privacy and scalability standpoint, we also suspect that, for any complex task requiring knowledge of a system's state where an appropriate simulation and execution environment is available for use in model training, using \dataname{} would almost certainly be less effective than a method that more fully leverages such information. In addition, while not all stages of \dataname{} are required to produce data, the results of Table \ref{res:bfcl_v3_abl_subtr} show the use of all stages provides the best performance. Thus, substituting a more effective frontier model for the open weights model used in our experiments, would result in a significantly more expensive cost per run than simpler approaches that execute in fewer stages (e.g., ToolAce-MT \cite{zeng2025toolace}).

%% file: sections/conclusion.tex
\section{Conclusions}

In this work we introduced a new method for producing complex, multi-turn tool calling interactions that does not rely on a stateful execution environment. Our approach contributed a novel selective generation strategy that allowed for a straightforward introduction of a back translation technique for noise mitigation. Using the BFCL \cite{patilberkeley}, $\tau$-bench \cite{yao2024tau}, and ACEBench \cite{chen2025acebench} benchmarks we showed that the novel aspects of our method contributed significantly to its performance, which exceeded that of much larger models and was competitive with other top performing tool calling approaches trained from more powerful teacher models. Lastly, we ensure all components of this work, from the data generation approach to the data itself, are open-sourced and available under a permissive license.

%% file: sections/appendix.tex
\section{Appendix}
\label{sec:appendix}

\subsection{Training Setup}

We trained our various base models using the LlamaFactory framework \cite{zheng2024llamafactory}. For all experiments, we used the same settings for LoRA \cite{hulora} fine-tuning as \cite{zeng2025toolace}, with a batch size of 64, a LoRA rank of 16, a learning rate of $1\mathrm{e}{-4}$, a cosine learning rate schedule, a warmup ratio of $0.1$, and 1 epoch of training. We used 8 H100 GPUs for our training run and 4 H100 GPUs to run the teacher model (\texttt{GPT-OSS-120B} for the main experiments).

\subsection{Additional Analyses}

\subsection{Cost}
\label{sec:cost_analysis}
For our main experiments used \texttt{GPT-OSS-120b}, which costs \$0.15 and \$0.60 per 1M input / output tokens, respectively\footnote{see https://azure.microsoft.com/en-us/pricing/details/azure-openai/\#pricing}. In comparison, ToolACE-MT \cite{liutoolace} uses GPT-4, which is priced at \$2.00 and \$8.00 per 1M input / output tokens. Our logging indicates that, with the Azure pricing, it costs about \$0.015 per example (factoring in rejected samples). Thus, to produce 5000 conversations it would cost about \$75.00. Our approach has not optimized for token efficiency, however, it would be an interesting direction to explore as there is certainly room for such optimizations, e.g., we use large sets of tool specifications during generation that account for millions of tokens and we do not prune away tools when they are not needed.

\subsection{Search and Non-Search Data}

An initial hypothesis we held was that implicit tool calls would almost always involve explicit search or information retrieval. To determine whether this was the case, we took the set of base data points and filtered out all conversations where the first tool call (which is always an implicit tool call) contained any of the following strings: "search", "list", "fetch", "retrieve", "get", "lookup", "enumerate", "find". To our surprise, this filtering left 43\% of the data (i.e., roughly 57\% of the data uses tool calls explicitly intended for retrieval). Of the remaining 43\% of conversations, the most common initial tool calls ($\sim30\%$) were those involving computing a number (e.g., \texttt{compute\_overall\_expense}, \texttt{compute\_mpg}, etc.). Following that, tool calls involving creation or generation were most common (e.g., \texttt{generate\_secure\_passcode}, \texttt{create\_email\_for\_user}, etc.).

Figure \ref{fig:non_search_example} provides an example of one such case. In the example, the user request involves producing a business card. As the first step in completing the request, the agent initializes a new card with the \texttt{create\_card} tool call \textit{without} being explicitly instructed to. While intuitive to a human, this behavior is challenging to extract from an LLM (particularly smaller LLMs).

\begin{figure}[t]
\begin{tcolorbox}[colback=gray!10,colframe=gray!50,rounded corners,boxrule=0.5pt,left=1mm, right=1mm, top=1mm, bottom=1mm]
\begin{tcolorbox}[colback=gray!10, colframe=gray!50, rounded corners, boxrule=0.5pt,left=1mm, right=1mm, top=1mm, bottom=1mm]
\begin{minted}[linenos=false,tabsize=4,fontsize=\scriptsize]{text}
[
 {
  "role": "user",
  "content": "My goal is to place the text \"John Doe\\nDevOps Engineer\\nQR: 
    https://example.com/johnqr\" at the bottom of a card that measures 3.5 inches wide by 2.0 inches 
    high."
 },
 {
  "role": "tool_calls",
  "content": [
   {
    "name": "create_card",
    "arguments": {
     "width": 3.5,
     "height": 2.0
    },
    "id": "$1"
   }
  ]
 },
 {
  "role": "tool",
  "content": {
   "card_id": "card_9f4b2c6e-8c3d-4a1f-9b2e-5d7f6a8c9b01"
  },
  "tool_call_id": "$1",
 },
 {
  "role": "tool_calls",
  "content": [
   {
    "name": "add_text_to_card",
    "arguments": {
     "card_id": "card_9f4b2c6e-8c3d-4a1f-9b2e-5d7f6a8c9b01",
     "text": "John Doe\nDevOps Engineer\nQR: https://example.com/johnqr",
     "position": "bottom"
    },
    "id": "$2"
   }
  ]
 },
 {
  "role": "tool",
  "content": {
   "card_id": "card_9f4b2c6e-8c3d-4a1f-9b2e-5d7f6a8c9b01",
   "updated_card": {
    "id": "card_9f4b2c6e-8c3d-4a1f-9b2e-5d7f6a8c9b01",
    "text": [
     "John Doe\nDevOps Engineer\nQR: https://example.com/johnqr"
    ]
   }
  },
  "tool_call_id": "$2",
 },
 {
  "role": "assistant",
  "content": "I created a ... card now contains that text."
 }
]
\end{minted}
\end{tcolorbox}
\end{tcolorbox}
\caption{Example of a conversation with implicit tool calls that does not use explicit search tool calls}
\label{fig:non_search_example}
\end{figure}

\subsection{Role of Kendall's $\tau$}

The usage of the Kendall rank correlation coefficient may at first seem redundant, as the prompt for generating the user request explicitly instructs the LLM which tool calls to treat as implicit. Interestingly, we found that without both the filter and the LLM instruction, conversations would often contain one of two clear issues. Without the rank correlation filter, the LLM would quite often just ignore the instruction and produce a conversation where every tool call was explicit. On the other hand, when just the rank filter was used and the prompt did not explicitly state which tool calls to treat as implicit, we found that the LLM would produce tool calls that did not reflect the underlying request. We provide examples of both of these error patterns in Figures \ref{fig:no_kendall_tau} and \ref{fig:kendall_tau_only}.

\begin{figure}[t]
\begin{tcolorbox}[colback=gray!10,colframe=gray!50,rounded corners,boxrule=0.5pt,left=1mm, right=1mm, top=1mm, bottom=1mm]
\begin{tcolorbox}[colback=gray!10, colframe=gray!50, rounded corners, boxrule=0.5pt,left=1mm, right=1mm, top=1mm, bottom=1mm]
\begin{minted}[linenos=false,tabsize=4,fontsize=\scriptsize]{text}
[
  {
    "role": "user",
    "content": "I need to assign the insurance policy identified by EMP-12345 to the TeacherAccess 
      role based on its rank. Please retrieve the details of that policy, determine its rank, and 
      attach the corresponding policy ARN to the TeacherAccess role."
    }
  },
  {
    "role": "assistant",
    "content": {
      "name": "getPolicyDetails",
      "arguments": { "policyNumber": "EMP-12345" },
      "id": "$1"
    }
  },
  {
    "role": "assistant",
    "content": {
      "name": "extract_policy_rank",
      "arguments": { "policy_details": "$1.policyDetails" },
      "id": "$2"
    }
  },
  {
    "role": "assistant",
    "content": {
      "name": "attach_policy_to_role",
      "arguments": { "role_name": "TeacherAccess", "policy_arn": "$2.policy_rank" },
      "id": "$3"
    }
  }
]
\end{minted}
\end{tcolorbox}
\end{tcolorbox}
\caption{Example where the LLM ignores the prompt and does not produce implicit tool calls}
\label{fig:no_kendall_tau}
\vskip 10pt
\begin{tcolorbox}[colback=gray!10,colframe=gray!50,rounded corners,boxrule=0.5pt,left=1mm, right=1mm, top=1mm, bottom=1mm]
\begin{tcolorbox}[colback=gray!10, colframe=gray!50, rounded corners, boxrule=0.5pt,left=1mm, right=1mm, top=1mm, bottom=1mm]
\begin{minted}[linenos=false,tabsize=4,fontsize=\scriptsize]{text}
[
  {
    "role": "user",
    "content": "I’m amazed that the latest post on Twitter disappears so quickly; now could you fetch 
      the new most recent tweet ID from the social_media_db by running the query `SELECT post_id FROM 
      tweets WHERE user_handle = 'alex_econ' ORDER BY created_at DESC LIMIT 1` with limit set to 1 
      and offset set to 0?"
  },
  {
    "role": "assistant",
    "content": {
      "name": "GetTestEndpointResponse",
      "arguments": {
        "database": "social_media_db",
        "query": "SELECT post_id FROM tweets WHERE user_handle = 'alex_econ' ORDER BY created_at 
          DESC LIMIT 1",
        "offset": 0,
        "limit": 1
      },
      "id": "$1"
    }
  },
  {
    "role": "assistant",
    "content": {
      "name": "deletePost",
      "arguments": { "platform": "Twitter", "post_id": "$1.result[0].column1" },
      "id": "$2"
    }
  }
]
\end{minted}
\end{tcolorbox}
\end{tcolorbox}
\caption{Example of a conversation that did not have the instruction to treat certain tool calls as implicit but that passes the rank correlation filter. Here the rank correlation filter is satisfied because the tool call does not accurately reflect the user's request}
\label{fig:kendall_tau_only}

\end{figure}

\clearpage

\subsection{Examples}

\subsubsection{Tool Specification}

\begin{figure}[t]
\begin{tcolorbox}[colback=gray!10,colframe=gray!50,rounded corners,boxrule=0.5pt,left=1mm, right=1mm, top=1mm, bottom=1mm]
\begin{tcolorbox}[colback=gray!10, colframe=gray!50, rounded corners, boxrule=0.5pt,left=1mm, right=1mm, top=1mm, bottom=1mm]
\begin{minted}[linenos=false,tabsize=4,fontsize=\scriptsize]{text}
{
  "search_cities": {
    "name": "search_cities",
    "description": "Search for cities ...",
    "parameters": {
      "properties": {
        "keyword": {
          "description": "The keyword ...",
          "type": "string"
        }
      },
    },
    "output_parameters": {
      "properties": {
        "cities": {
          "type": "array",
          "items": {
            "type": "object",
            "properties": {
              "name": {
                "type": "string"
              },
              "country": {
                "type": "string"
              },
            }
        ...
    ...
}
\end{minted}
\end{tcolorbox}
\end{tcolorbox}
\caption{Example tool specification}
\label{fig:tool_spec}
\end{figure}

\subsubsection{Full Conversation}

\begin{figure*}
\begin{tcolorbox}[
        colback=gray!5, 
        colframe=gray!70, 
        rounded corners,
        boxrule=1pt, 
        left=1mm, right=1mm, top=1mm, bottom=1mm
    ]
\begin{tcolorbox}[colback=gray!10, colframe=gray!50, rounded corners, boxrule=0.5pt,left=1mm, right=1mm, top=1mm, bottom=1mm]
\begin{minted}[breaklines,linenos=false,tabsize=4,fontsize=\tiny]{text}
[
    {
        "role": "user",
        "content": "I want you to figure out the carbon footprint of my refrigerator when it runs on HydroElectric power, using a 150\u2011watt rating and operating for 24 hours each day."
    },
    {
        "role": "tool_calls",
        "content": [
            {
                "name": "calculatePowerConsumption",
                "arguments": {
                    "device_name": "refrigerator",
                    "power_rating": 150,
                    "usage_time": 24
                },
                "id": "$1"
            }
        ]
    },
    {
        "role": "tool",
        "content": {
            "power_consumption": 3.6
        },
        "tool_call_id": "$1",
        "type": "json"
    },
    {
        "role": "tool_calls",
        "content": [
            {
                "name": "CalculateCarbonFootprint",
                "arguments": {
                    "energy": "HydroElectric",
                    "consumption": 3.6
                },
                "id": "$2"
            }
        ]
    },
    {
        "role": "tool",
        "content": {
            "carbon_footprint": 0.0144
        },
        "tool_call_id": "$2",
        "type": "json"
    },
    {
        "role": "assistant",
        "content": "I calculated that my refrigerator consumes 3.6\u202fkWh per day, resulting in a carbon footprint of about 0.0144\u202fkg CO\u2082e when powered by hydroelectric energy."
    }
]
\end{minted}
\end{tcolorbox}
\end{tcolorbox} 
\caption{Example of full interaction}
\label{fig:full_interaction}
\end{figure*}

\clearpage

\subsection{Prompts}

\begin{figure*}
\begin{tcolorbox}[
        colback=gray!5, 
        colframe=gray!70, 
        rounded corners,
        boxrule=1pt, 
        left=1mm, right=1mm, top=1mm, bottom=1mm
    ]
\begin{tcolorbox}[colback=gray!10, colframe=gray!50, rounded corners, boxrule=0.5pt,left=1mm, right=1mm, top=1mm, bottom=1mm]
\begin{minted}[breaklines,linenos=false,tabsize=4,fontsize=\tiny]{text}
You are a tool chaining agent

You will be provided with:
    - The set of tools the agent has access to between the <tools> and </tools> tags
{% if persona %}    - A user persona that is the individual who has made the tool calls between the <persona> and </persona> tags{% endif %}
    - A description of the person who has made the tool calls in between the <details> and </details> tags
    - The current date will be provided in between the <date> and </date> tags
    - A target count that represents the maximum number of tool calls you should generate in between the <length> and </length> tags
    - A list of previously made tool calls in between the <prior_tool_calls> and </prior_tool_calls> tags

Your task is to simulate a chained sequence of tool calls in between the <tool_calls> and </tool_calls> tags that makes use of the tools from <tools>. The output format should be 

<tool_calls>
[
    {
        "name": name of first tool call
        "arguments": {
            "parameter name": parameter value
            ...
        }
        "id": "$1"
    },
    {
        "name": name of second tool call
        "arguments": {
            "parameter name": parameter value
            ...
        }
        "id": "$2"
    },
    ...
]
</tool_calls>

{% if persona %}The tool calls you simulate should be as if they were written by the individual specified by the <persona>{% endif %}
The arguments to the tool calls should be influenced by the details in <details>.

On each "assistant" turn, you should produce a list of tool calls (at most as many as specified in <length>) in between the <tool_calls> and </tool_calls> tags that make use of the provided tools.
{% if allow_nested %}The MOST IMPORTANT requirement is that you should ensure that each tool call after the first has one or more arguments in "arguments" that reference an "output_parameter" of an earlier tool call (either from <tool_calls> or <prior_tool_calls>) using pythonic dot notation, e.g., "$1.output_parameter_here"{% endif %}

The messages you generate must meet the following criteria:
    - Each "assistant" tool call must have a "id" that is "$" + its position in the list, e.g., "$1", "$2", "$3", etc.
    - Your requests should involve SEVERAL of the "arguments" to each tool that is required, i.e., make each tool call complex
    - You can generate up to as many "assistant" tool calls as specified by <length> though you can generate fewer
    - Make sure when you use dates and times in your tool calls, that you copy them as-is from the user request (i.e., do NOT change the date or time formatting)
    - Use realistic values for the arguments to each tool call rather than hypothetical or placeholder values
    - Make sure each tool call has several arguments
    {% if allow_nested %}- Try to make sure the tool calls are connected, i.e., make tool calls that have an argument in "arguments" that references an "output_parameter" of another tool call using pythonic dot notation, e.g., "$1.output_parameter_here"{% endif -%}
    - Do NOT reference input "parameter" fields from "arguments" with pythonic dot notation (only "output_parameter" fields)
    - If there are tool calls in <prior_tool_calls>, the "id" of tool calls should pick up from the last "id" in <prior_tool_calls>
    {% if allow_nested %}- If there are tool calls in <prior_tool_calls>, at least one tool call has an argument in "arguments" that references an "output_parameter" of one of the tool calls in <prior_tool_calls>{% endif -%}
    - In your list of tool calls, do NOT use the same tools that are used in <prior_tool_calls>
\end{minted}
\end{tcolorbox}
\end{tcolorbox} 
\caption{Prompt for Plan step}
\label{fig:prompt_plan}
\end{figure*}

\begin{figure*}
\begin{tcolorbox}[
        colback=gray!5, 
        colframe=gray!70, 
        rounded corners,
        boxrule=1pt, 
        left=1mm, right=1mm, top=1mm, bottom=1mm
    ]
\begin{tcolorbox}[colback=gray!10, colframe=gray!50, rounded corners, boxrule=0.5pt,left=1mm, right=1mm, top=1mm, bottom=1mm]
\begin{minted}[breaklines,linenos=false,tabsize=4,fontsize=\tiny]{text}
You will be provided with:
    - The set of tools the agent has access to between the <tools> and </tools> tags
    - A list of "required" elements your user utterance MUST include in between the <required> and </required> tags
    - A user persona that is the individual who has made the tool calls between the <persona> and </persona> tags
    - The conversation history will be given between the <history> and </history> tags, and will consist of "user" requests followed by the "assistant" tool calls
    - Any hidden tool calls that are invisible to the "user" but must executed by the system to help solve the "user" request will be given between the <hidden> and </hidden> tags
    - The list of explicit tool calls that was the original goal of the "user" in between the <explicit> and </explicit> tags

Your task is to predict what the "user" request in between the <request> and </request> tags.

The "user" request you generate must meet the following criteria:
    - Each "user" request either does not reference the assistant or only refers to the assistant as "you"
    - The user request you predict should be as if it was written by the individual specified by the <persona>
    - If <history> is not empty, make sure the "user" request is a follow-up question to the last "user" message in the list by indirectly referencing values from prior tool calls in <history>
    - Make sure to write the request in between the <request> and </request> tags
    - Write the "user" request as if the "user" was {{ user_attribute }}
    - You must naturally incorporate each element of <required> in your "user" request, i.e., if an element is a list or dictionary, naturally incorporate its elements (don't just copy over the raw object itself)
    - Always state the "user" request in the present tense, i.e., there should not be any words that are in the past tense
    - The "user" request should instruct the "assistant" to do something
    - The "user" request should NEVER describe any tool call from <hidden>
    - The "user" request should NEVER directly reference tool call IDs, tool names, tool parameters, or tool descriptions
    - The "user" request should start by addressing the goals of the user, which is the tool call sequence in <explicit>
\end{minted}
\end{tcolorbox}
\end{tcolorbox} 
\caption{Prompt for Request Generation step}
\label{fig:prompt_req_gen}
\end{figure*}

\begin{figure*}
\begin{tcolorbox}[
        colback=gray!5, 
        colframe=gray!70, 
        rounded corners,
        boxrule=1pt, 
        left=1mm, right=1mm, top=1mm, bottom=1mm
    ]
\begin{tcolorbox}[colback=gray!10, colframe=gray!50, rounded corners, boxrule=0.5pt,left=1mm, right=1mm, top=1mm, bottom=1mm]
\begin{minted}[breaklines,linenos=false,tabsize=4,fontsize=\tiny]{text}
You are a agent that responds to user requests with tool calls

You will be provided with:
    - The set of tools the agent has access to between the <tools> and </tools> tags
    - The conversation history will be given between the <history> and </history> tags, and will consist of "user" requests followed by the "assistant" tool calls
    - A list of "required" elements your tool call list MUST include in between the <required> and </required> tags
    - Any "user" request to produce the tool calls for in between the <request> and </request> tags

Your task is to produce the next list of tool calls for the user's <request> in between the <tool_calls> and </tool_calls> tags that makes use of tools from <tools>. The output format should be 

<tool_calls>
[
    {
        "name": name of first tool call
        "arguments": {
            "parameter name": parameter value
            ...
        }
        "id": "$1"
    },
    {
        "name": name of second tool call
        "arguments": {
            "parameter name": parameter value
            ...
        }
        "id": "$2"
    },
    ...
]
</tool_calls>

The messages you generate must meet the following criteria:
    - Each tool call must have a "id" that is "$" + its position in the list, e.g., "$1", "$2", "$3", etc.
    - The tool call ids should continue across the whole conversation
    - Everything you generate MUST use valid UTF-8 characters
    - Your requests should involve SEVERAL of the "arguments" to each tool that is required, i.e., make each tool call complex
    - Make sure when you use dates and times in your tool calls, that you copy them as-is from the user request (i.e., do NOT change the date or time formatting)
    - The arguments of tool calls in each "assistant" message MUST be taken from the "user" requests or the output will be FAILED
    - You must use each element of <required> as arguments to the tool calls
    - Tool calls must be used across "user" utterances (e.g., a tool call result from the first "user" utterance used for the third "user" utterance). This should be done with the dot notation, e.g., "$1.output_parameter_here"
    - If <history> is not empty, make sure the "id" of tool calls should pick up from the last "id" of a "assistant" turn in the list
    - If <history> is not empty, at least one tool call has an argument in "arguments" that references an "output_parameter" of one of the "assistant" tool calls in <history>
\end{minted}
\end{tcolorbox}
\end{tcolorbox} 
\caption{Prompt for Back Translation step}
\label{fig:prompt_bt}
\end{figure*}

\begin{figure*}
\begin{tcolorbox}[
        colback=gray!5, 
        colframe=gray!70, 
        rounded corners,
        boxrule=1pt, 
        left=1mm, right=1mm, top=1mm, bottom=1mm
    ]
\begin{tcolorbox}[colback=gray!10, colframe=gray!50, rounded corners, boxrule=0.5pt,left=1mm, right=1mm, top=1mm, bottom=1mm]
\begin{minted}[breaklines,linenos=false,tabsize=4,fontsize=\tiny]{text}
You are a tutoring agent that provides hints to a student that will help them correct a mistake they've made when writing a tool call

You will be provided with:
    - The set of tools the student has access to between the <tools> and </tools> tags
    - The expected, correct tool call that should have been received between the <correct_tool_call> and </correct_tool_call> tags
    - The student's tool call that is wrong and needs to be corrected between the <incorrect_tool_call> and </incorrect_tool_call> tags

Your task is to analyze the difference between the correct tool call and the correct tool call (with respect to the tools in <tools>) and provide a subtle hint to the student as to what they should fix.
You MUST make your hint compact and concise.

The hint you write in between the <hint> and </hint> tags should obey the following guidelines:
    - The hint does not directly mention any name or parameter of a tool from <tools>
    - If the tool name is wrong or the arguments are wrong, use the "description" of the correct tool name or correct arguments to point out what the student should fix
    - Identify any missing values or extra values provided in the incorrect tool call
    - If the tool is different, guide the student to the correct tool without explicitly using the name of the tool
    - Be as concise as possible with your hint and use no more than 10 words
\end{minted}
\end{tcolorbox}
\end{tcolorbox} 
\caption{Prompt for Error Hint step}
\label{fig:prompt_err_hint}
\end{figure*}

\begin{figure*}
\begin{tcolorbox}[
        colback=gray!5, 
        colframe=gray!70, 
        rounded corners,
        boxrule=1pt, 
        left=1mm, right=1mm, top=1mm, bottom=1mm
    ]
\begin{tcolorbox}[colback=gray!10, colframe=gray!50, rounded corners, boxrule=0.5pt,left=1mm, right=1mm, top=1mm, bottom=1mm]
\begin{minted}[breaklines,linenos=false,tabsize=4,fontsize=\tiny]{text}
You are an agent that detects missing tools

You will be provided with:
    - A "user" request in between the <user> and </user> tags

Your task is to produce a concise, clear "assistant" statement in between the <statement> and </statement> tags indicating that the available tools are insufficient for the "user" request.

The new response you generate must meet the following criteria:
    - Everything you generate MUST use valid UTF-8 characters
    - Your "assistant" statement should be written in between <statement> and </statement> 
    - Keep the "assistant" statement brief and do not mention specifics, i.e., just indicate that the request cannot be completed
\end{minted}
\end{tcolorbox}
\end{tcolorbox} 
\caption{Prompt for Missing Function clarification step}
\label{fig:prompt_miss_func}
\end{figure*}

\begin{figure*}
\begin{tcolorbox}[
        colback=gray!5, 
        colframe=gray!70, 
        rounded corners,
        boxrule=1pt, 
        left=1mm, right=1mm, top=1mm, bottom=1mm
    ]
\begin{tcolorbox}[colback=gray!10, colframe=gray!50, rounded corners, boxrule=0.5pt,left=1mm, right=1mm, top=1mm, bottom=1mm]
\begin{minted}[breaklines,linenos=false,tabsize=4,fontsize=\tiny]{text}
You are an agent that asks an unsatisfiable user request

You will be provided with:
    - The set of tools the agent has access to between the <tools> and </tools> tags
    - A complete "user" request in between the <user> and </user> tags
    - The tool calls that the agent performs to satisfy the user request in between the <tool_calls> and </tool_calls> tags
    - A list of terms used as arguments in the tool calls to exclude in between the <exclude> and </exclude> tags

Your task is to produce a new "user" request that is a more incomplete version of the "user" request in <user> that does not use any of the arguments in <exclude>
After that, write a concise, polite follow-up "assistant" clarification request that asks the "user" for more details regarding the tool parameters associated with the missing arguments based on the new request.
To finish, produce an answer from the "user" that provides the details that the "assistant" asked for about the missing arguments.
You should write the "user" request in between the <request> and </request> tags, the "assistant" clarification in between the <clarification> and </clarification> tags, and the "user" answer between the <answer> and </answer> tags.

The new request you generate must meet the following criteria:
    - Everything you generate MUST use valid UTF-8 characters and in English.
    - Your "user" request should almost be the same as the request in <user> and </user> but missing only the information from <exclude>
    - Your "user" request should be written in between <request> and </request>
    - Your "user" request should NOT request any information that can be found in the tool outputs
    - Your "assistant" request should be written in between <clarification> and </clarification> and should be brief while NOT directly referencing anything in <tool_calls>
    - Your "user" answer should be direct, answering the "assistant" clarification question and must use the information in the <tool_calls> to supply the missing parameter
    - You "user" answer should be written in between <answer> and </answer> 
\end{minted}
\end{tcolorbox}
\end{tcolorbox} 
\caption{Prompt for Missing Parameter clarification step}
\label{fig:prompt_miss_param}
\end{figure*}

\begin{figure*}
\begin{tcolorbox}[
        colback=gray!5, 
        colframe=gray!70, 
        rounded corners,
        boxrule=1pt, 
        left=1mm, right=1mm, top=1mm, bottom=1mm
    ]
\begin{tcolorbox}[colback=gray!10, colframe=gray!50, rounded corners, boxrule=0.5pt,left=1mm, right=1mm, top=1mm, bottom=1mm]
\begin{minted}[breaklines,linenos=false,tabsize=4,fontsize=\tiny]{text}
As a data quality evaluator, you must assess a conversation for the alignment between its user queries, corresponding tool calls, and tool call execution execution results.
These function calls and results are generated by other models, and your task is to ensure these results accurately reflect the user's intentions.

Do not pass if:
    - Any function call contains arguments that are hallucinated or placeholders
    - For a user request, if it refers to any of the upcoming tool call outputs in past tense

You will be given:
    - The list of all available functions in between the <tools> and </tools> tags
    - The complete conversation between the <conversation> and </conversation> tags

Note: The query may have multiple intentions. Functions may be placeholders, and execution results may be truncated due to length, which is acceptable and should not cause a failure.
The main decision factor is whether the function calls accurately reflect the query's intentions and the function descriptions.

{% if reasoning %}Provide your reasoning in between the <thought> and </thought> tags.
{% endif %}Then write your deliberation as to whether or not the data passes (answer yes or no) in between the <pass> and </pass> tags.

Your response MUST strictly adhere to the following format, and NO other text MUST be included:

{% if reasoning %}<thought>
Concisely describe your reasoning here
</thought>
{% endif %}
<pass>
"yes" or "no"
</pass>
\end{minted}
\end{tcolorbox}
\end{tcolorbox} 
\caption{Prompt for LLM-as-a-judge filtering}
\label{fig:prompt_filter}
\end{figure*}

%% file: sections/checklist.tex
\section*{NeurIPS Paper Checklist}

The checklist is designed to encourage best practices for responsible machine learning research, addressing issues of reproducibility, transparency, research ethics, and societal impact. Do not remove the checklist: {\bf The papers not including the checklist will be desk rejected.} The checklist should follow the references and follow the (optional) supplemental material.  The checklist does NOT count towards the page
limit. 

Please read the checklist guidelines carefully for information on how to answer these questions. For each question in the checklist:
\begin{itemize}
    \item You should answer \answerYes{}, \answerNo{}, or \answerNA{}.
    \item \answerNA{} means either that the question is Not Applicable for that particular paper or the relevant information is Not Available.
    \item Please provide a short (1--2 sentence) justification right after your answer (even for \answerNA). 
\end{itemize}

{\bf The checklist answers are an integral part of your paper submission.} They are visible to the reviewers, area chairs, senior area chairs, and ethics reviewers. You will also be asked to include it (after eventual revisions) with the final version of your paper, and its final version will be published with the paper.

The reviewers of your paper will be asked to use the checklist as one of the factors in their evaluation. While \answerYes{} is generally preferable to \answerNo{}, it is perfectly acceptable to answer \answerNo{} provided a proper justification is given (e.g., error bars are not reported because it would be too computationally expensive'' or ``we were unable to find the license for the dataset we used''). In general, answering \answerNo{} or \answerNA{} is not grounds for rejection. While the questions are phrased in a binary way, we acknowledge that the true answer is often more nuanced, so please just use your best judgment and write a justification to elaborate. All supporting evidence can appear either in the main paper or the supplemental material, provided in appendix. If you answer \answerYes{} to a question, in the justification please point to the section(s) where related material for the question can be found.

IMPORTANT, please:
\begin{itemize}
    \item {\bf Delete this instruction block, but keep the section heading ``NeurIPS Paper Checklist"},
    \item  {\bf Keep the checklist subsection headings, questions/answers and guidelines below.}
    \item {\bf Do not modify the questions and only use the provided macros for your answers}.
\end{itemize}


\begin{enumerate}

\item {\bf Claims}
    \item[] Question: Do the main claims made in the abstract and introduction accurately reflect the paper's contributions and scope?
    \item[] Answer: \answerYes{} 
    \item[] Justification: We clearly state our contributions and scope in the introduction (repeating them at the end of the introduction for emphasis)

\item {\bf Limitations}
    \item[] Question: Does the paper discuss the limitations of the work performed by the authors?
    \item[] Answer: \answerYes{} 
    \item[] Justification: We have a dedicated limitations section

\item {\bf Theory assumptions and proofs}
    \item[] Question: For each theoretical result, does the paper provide the full set of assumptions and a complete (and correct) proof?
    \item[] Answer: \answerNA{} 
    \item[] Justification: The paper does not include theoretical results

    \item {\bf Experimental result reproducibility}
    \item[] Question: Does the paper fully disclose all the information needed to reproduce the main experimental results of the paper to the extent that it affects the main claims and/or conclusions of the paper (regardless of whether the code and data are provided or not)?
    \item[] Answer: \answerYes{} 
    \item[] Justification: We have provided all the implementation details as well as prompts used to generate data. We also provide code as supplementary material

\item {\bf Open access to data and code}
    \item[] Question: Does the paper provide open access to the data and code, with sufficient instructions to faithfully reproduce the main experimental results, as described in supplemental material?
    \item[] Answer: \answerYes{} 
    \item[] Justification: We provide open access to the data and code, as well as the exact training sets used to train the model used for evaluation

\item {\bf Experimental setting/details}
    \item[] Question: Does the paper specify all the training and test details (e.g., data splits, hyperparameters, how they were chosen, type of optimizer) necessary to understand the results?
    \item[] Answer: \answerYes{} 
    \item[] Justification: The full details are provided in the appendix and the provided code

\item {\bf Experiment statistical significance}
    \item[] Question: Does the paper report error bars suitably and correctly defined or other appropriate information about the statistical significance of the experiments?
    \item[] Answer: \answerYes{} 
    \item[] Justification: Where applicable, those numbers are provided

\item {\bf Experiments compute resources}
    \item[] Question: For each experiment, does the paper provide sufficient information on the computer resources (type of compute workers, memory, time of execution) needed to reproduce the experiments?
    \item[] Answer: \answerYes{} 
    \item[] Justification: We have provided such information in the appendix
    
\item {\bf Code of ethics}
    \item[] Question: Does the research conducted in the paper conform, in every respect, with the NeurIPS Code of Ethics \url{https://neurips.cc/public/EthicsGuidelines}?
    \item[] Answer: \answerYes{} 
    \item[] Justification: The research conforms to the ethics code

\item {\bf Broader impacts}
    \item[] Question: Does the paper discuss both potential positive societal impacts and negative societal impacts of the work performed?
    \item[] Answer: \answerYes{} 
    \item[] Justification: The impacts are justified in the introduction
    
\item {\bf Safeguards}
    \item[] Question: Does the paper describe safeguards that have been put in place for responsible release of data or models that have a high risk for misuse (e.g., pre-trained language models, image generators, or scraped datasets)?
    \item[] Answer: \answerNA{} 
    \item[] Justification: We do not believe our work poses such risks

\item {\bf Licenses for existing assets}
    \item[] Question: Are the creators or original owners of assets (e.g., code, data, models), used in the paper, properly credited and are the license and terms of use explicitly mentioned and properly respected?
    \item[] Answer: \answerYes{} 
    \item[] Justification: All works by other creators are properly credited and their licenses respected

\item {\bf New assets}
    \item[] Question: Are new assets introduced in the paper well documented and is the documentation provided alongside the assets?
    \item[] Answer: \answerYes{}{} 
    \item[] Justification: The code and data are well documented in their release

\item {\bf Crowdsourcing and research with human subjects}
    \item[] Question: For crowdsourcing experiments and research with human subjects, does the paper include the full text of instructions given to participants and screenshots, if applicable, as well as details about compensation (if any)? 
    \item[] Answer: \answerNA{} 
    \item[] Justification: the paper does not involve crowdsourcing nor research with human subjects

\item {\bf Institutional review board (IRB) approvals or equivalent for research with human subjects}
    \item[] Question: Does the paper describe potential risks incurred by study participants, whether such risks were disclosed to the subjects, and whether Institutional Review Board (IRB) approvals (or an equivalent approval/review based on the requirements of your country or institution) were obtained?
    \item[] Answer: \answerNA{} 
    \item[] Justification: The paper does not involve crowdsourcing nor research with human subjects

\item {\bf Declaration of LLM usage}
    \item[] Question: Does the paper describe the usage of LLMs if it is an important, original, or non-standard component of the core methods in this research? Note that if the LLM is used only for writing, editing, or formatting purposes and does \emph{not} impact the core methodology, scientific rigor, or originality of the research, declaration is not required.
    \item[] Answer: \answerNA{} 
    \item[] Justification: The core method development in this research does not involve LLMs as non-standard components

\end{enumerate}

%% file: main.bbl
\begin{thebibliography}{51}
\providecommand{\natexlab}[1]{#1}
\providecommand{\url}[1]{\texttt{#1}}
\expandafter\ifx\csname urlstyle\endcsname\relax
  \providecommand{\doi}[1]{doi: #1}\else
  \providecommand{\doi}{doi: \begingroup \urlstyle{rm}\Url}\fi

\bibitem[AI@Meta(2024)]{llama-3}
AI@Meta.
\newblock Llama 3 model card.
\newblock 2024.
\newblock URL
  \url{https://github.com/meta-llama/llama3/blob/main/MODEL_CARD.md}.

\bibitem[Basu et~al.(2025)Basu, Abdelaziz, Kate, Agarwal, Crouse, Rizk,
  Bradford, Munawar, Kumaravel, Goyal, Wang, Lastras, and
  Kapanipathi]{basu2025nestfulbenchmarkevaluatingllms}
Kinjal Basu, Ibrahim Abdelaziz, Kiran Kate, Mayank Agarwal, Maxwell Crouse,
  Yara Rizk, Kelsey Bradford, Asim Munawar, Sadhana Kumaravel, Saurabh Goyal,
  Xin Wang, Luis~A. Lastras, and Pavan Kapanipathi.
\newblock {NESTFUL}: A benchmark for evaluating {LLM}s on nested sequences of
  {API} calls.
\newblock In Christos Christodoulopoulos, Tanmoy Chakraborty, Carolyn Rose, and
  Violet Peng, editors, \emph{Proceedings of the 2025 Conference on Empirical
  Methods in Natural Language Processing}, pages 33526--33535, Suzhou, China,
  November 2025. Association for Computational Linguistics.
\newblock ISBN 979-8-89176-332-6.
\newblock \doi{10.18653/v1/2025.emnlp-main.1702}.
\newblock URL \url{https://aclanthology.org/2025.emnlp-main.1702/}.

\bibitem[Belcak et~al.(2025)Belcak, Heinrich, Diao, Fu, Dong, Muralidharan,
  Lin, and Molchanov]{belcak2025smalllanguagemodelsfuture}
Peter Belcak, Greg Heinrich, Shizhe Diao, Yonggan Fu, Xin Dong, Saurav
  Muralidharan, Yingyan~Celine Lin, and Pavlo Molchanov.
\newblock Small language models are the future of agentic ai, 2025.
\newblock URL \url{https://arxiv.org/abs/2506.02153}.

\bibitem[Bi et~al.(2025)Bi, Chen, Tseng, Zhang, Wang, Luo, Chen, Huang, Guan,
  Hao, et~al.]{bi2025gpt}
Ziqian Bi, Keyu Chen, Chiung-Yi Tseng, Danyang Zhang, Tianyang Wang, Hongying
  Luo, Lu~Chen, Junming Huang, Jibin Guan, Junfeng Hao, et~al.
\newblock Is gpt-oss good? a comprehensive evaluation of openai's latest open
  source models.
\newblock \emph{arXiv preprint arXiv:2508.12461}, 2025.

\bibitem[Chen et~al.(2025{\natexlab{a}})Chen, Hao, Liu, Huang, Zeng, Yu, Li,
  Wang, Gan, Huang, et~al.]{chen2025acebench}
Chen Chen, Xinlong Hao, Weiwen Liu, Xu~Huang, Xingshan Zeng, Shuai Yu, Dexun
  Li, Shuai Wang, Weinan Gan, Yuefeng Huang, et~al.
\newblock Acebench: Who wins the match point in tool usage?
\newblock \emph{arXiv preprint arXiv:2501.12851}, 2025{\natexlab{a}}.

\bibitem[Chen et~al.(2025{\natexlab{b}})Chen, Zhou, Lin, Feng, Shen, and
  Wu]{chen2025survey}
Kang Chen, Xiuze Zhou, Yuanguo Lin, Shibo Feng, Li~Shen, and Pengcheng Wu.
\newblock A survey on privacy risks and protection in large language models.
\newblock \emph{Journal of King Saud University Computer and Information
  Sciences}, 37\penalty0 (7):\penalty0 163, 2025{\natexlab{b}}.

\bibitem[Chen et~al.()Chen, Zaharia, and Zou]{chenfrugalgpt}
Lingjiao Chen, Matei Zaharia, and James Zou.
\newblock Frugalgpt: How to use large language models while reducing cost and
  improving performance.
\newblock \emph{Transactions on Machine Learning Research}.

\bibitem[Chen et~al.(2024)Chen, Sun, Li, Yang, Liang, Lu, Cui, Zhang, Zhou, and
  Chen]{chen2024facilitating}
Mingyang Chen, Haoze Sun, Tianpeng Li, Fan Yang, Hao Liang, Keer Lu, Bin Cui,
  Wentao Zhang, Zenan Zhou, and Weipeng Chen.
\newblock Facilitating multi-turn function calling for llms via compositional
  instruction tuning.
\newblock \emph{arXiv preprint arXiv:2410.12952}, 2024.

\bibitem[Desai et~al.(2024)Desai, Ravi, Luqman, Mallya, Kota, and
  Yadav]{desai2024opportunities}
Akshar~Prabhu Desai, Tejasvi Ravi, Mohammad Luqman, Ganesh Mallya, Nithya Kota,
  and Pranjul Yadav.
\newblock Opportunities and challenges of generative-ai in finance.
\newblock In \emph{2024 IEEE International Conference on Big Data (BigData)},
  pages 4913--4920. IEEE, 2024.

\bibitem[Farn and Shin(2023)]{farn2023tooltalk}
Nicholas Farn and Richard Shin.
\newblock Tooltalk: Evaluating tool-usage in a conversation setting.
\newblock \emph{arXiv preprint arXiv:2311.10775}, 2023.

\bibitem[Guo et~al.(2024)Guo, Cheng, Wang, Liang, Qin, Li, Liu, Sun, and
  Liu]{guo2024stabletoolbench}
Zhicheng Guo, Sijie Cheng, Hao Wang, Shihao Liang, Yujia Qin, Peng Li, Zhiyuan
  Liu, Maosong Sun, and Yang Liu.
\newblock Stabletoolbench: Towards stable large-scale benchmarking on tool
  learning of large language models.
\newblock \emph{arXiv preprint arXiv:2403.07714}, 2024.

\bibitem[Hu et~al.(2022)Hu, Wallis, Allen-Zhu, Li, Wang, Wang, Chen,
  et~al.]{hulora}
Edward~J Hu, Phillip Wallis, Zeyuan Allen-Zhu, Yuanzhi Li, Shean Wang, Lu~Wang,
  Weizhu Chen, et~al.
\newblock Lora: Low-rank adaptation of large language models.
\newblock In \emph{International Conference on Learning Representations}, 2022.

\bibitem[Hurst et~al.(2024)Hurst, Lerer, Goucher, Perelman, Ramesh, Clark,
  Ostrow, Welihinda, Hayes, Radford, et~al.]{hurst2024gpt}
Aaron Hurst, Adam Lerer, Adam~P Goucher, Adam Perelman, Aditya Ramesh, Aidan
  Clark, AJ~Ostrow, Akila Welihinda, Alan Hayes, Alec Radford, et~al.
\newblock Gpt-4o system card.
\newblock \emph{arXiv preprint arXiv:2410.21276}, 2024.

\bibitem[{IBM Research}(2026)]{granite2026}
{IBM Research}.
\newblock Granite 4.1 language models.
\newblock \url{https://huggingface.co/blog/ibm-granite/granit-4-1}, 2026.
\newblock Accessed: 2026-04-28.

\bibitem[Iqbal et~al.(2024)Iqbal, Kohno, and Roesner]{iqbal2024llm}
Umar Iqbal, Tadayoshi Kohno, and Franziska Roesner.
\newblock Llm platform security: Applying a systematic evaluation framework to
  openai's chatgpt plugins.
\newblock In \emph{Proceedings of the AAAI/ACM Conference on AI, Ethics, and
  Society}, volume~7, pages 611--623, 2024.

\bibitem[Jimenez~Gutierrez et~al.(2022)Jimenez~Gutierrez, McNeal, Washington,
  Chen, Li, Sun, and Su]{jimenez-gutierrez-etal-2022-thinking}
Bernal Jimenez~Gutierrez, Nikolas McNeal, Clayton Washington, You Chen, Lang
  Li, Huan Sun, and Yu~Su.
\newblock Thinking about {GPT}-3 in-context learning for biomedical {IE}? think
  again.
\newblock In Yoav Goldberg, Zornitsa Kozareva, and Yue Zhang, editors,
  \emph{Findings of the Association for Computational Linguistics: EMNLP 2022},
  pages 4497--4512, Abu Dhabi, United Arab Emirates, December 2022. Association
  for Computational Linguistics.
\newblock \doi{10.18653/v1/2022.findings-emnlp.329}.
\newblock URL \url{https://aclanthology.org/2022.findings-emnlp.329/}.

\bibitem[Khandelwal et~al.(2026)Khandelwal, Punnavajhala, Bhargav, Pandey,
  Joshi, Karanam, and Raghu]{khandelwal2026toolweave}
Dinesh Khandelwal, Gnana~Prakash Punnavajhala, G~P~Shrivatsa Bhargav, Gaurav
  Pandey, Sachindra Joshi, Hima Karanam, and Dinesh Raghu.
\newblock {TOOLWEAVE}: {FINE}-{GRAINED} {AND} {CONTROLLABLE} {SYNTHETIC} {DATA}
  {GENERATION} {FOR} {MULTI}-{TURN} {TOOL} {CALLING} {WITH} {NON}-{FRONTIER}
  {LLMS}, 2026.
\newblock URL \url{https://openreview.net/forum?id=KM7qycd8EB}.

\bibitem[Li et~al.(2023)Li, Zhao, Yu, Song, Li, Yu, Li, Huang, and
  Li]{li-etal-2023-api}
Minghao Li, Yingxiu Zhao, Bowen Yu, Feifan Song, Hangyu Li, Haiyang Yu, Zhoujun
  Li, Fei Huang, and Yongbin Li.
\newblock {API}-bank: A comprehensive benchmark for tool-augmented {LLM}s.
\newblock In Houda Bouamor, Juan Pino, and Kalika Bali, editors,
  \emph{Proceedings of the 2023 Conference on Empirical Methods in Natural
  Language Processing}, pages 3102--3116, Singapore, December 2023. Association
  for Computational Linguistics.
\newblock \doi{10.18653/v1/2023.emnlp-main.187}.
\newblock URL \url{https://aclanthology.org/2023.emnlp-main.187/}.

\bibitem[Liu et~al.(2024{\natexlab{a}})Liu, Feng, Xue, Wang, Wu, Lu, Zhao,
  Deng, Zhang, Ruan, et~al.]{liu2024deepseek}
Aixin Liu, Bei Feng, Bing Xue, Bingxuan Wang, Bochao Wu, Chengda Lu, Chenggang
  Zhao, Chengqi Deng, Chenyu Zhang, Chong Ruan, et~al.
\newblock Deepseek-v3 technical report.
\newblock \emph{arXiv preprint arXiv:2412.19437}, 2024{\natexlab{a}}.

\bibitem[Liu et~al.(2024{\natexlab{b}})Liu, Huang, Zeng, Yu, Li, Wang, Gan,
  Liu, Yu, WANG, et~al.]{liutoolace}
Weiwen Liu, Xu~Huang, Xingshan Zeng, Shuai Yu, Dexun Li, Shuai Wang, Weinan
  Gan, Zhengying Liu, Yuanqing Yu, Zezhong WANG, et~al.
\newblock Toolace: Winning the points of llm function calling.
\newblock In \emph{The Thirteenth International Conference on Learning
  Representations}, 2024{\natexlab{b}}.

\bibitem[Liu et~al.(2024{\natexlab{c}})Liu, Hoang, Zhang, Zhu, Lan, Tan, Yao,
  Liu, Feng, RN, et~al.]{liu2024apigen}
Zuxin Liu, Thai Hoang, Jianguo Zhang, Ming Zhu, Tian Lan, Juntao Tan, Weiran
  Yao, Zhiwei Liu, Yihao Feng, Rithesh RN, et~al.
\newblock Apigen: Automated pipeline for generating verifiable and diverse
  function-calling datasets.
\newblock \emph{Advances in Neural Information Processing Systems},
  37:\penalty0 54463--54482, 2024{\natexlab{c}}.

\bibitem[Lu et~al.(2025)Lu, Holleis, Zhang, Aumayer, Nan, Bai, Ma, Ma, Li, Yin,
  Wang, and Pang]{lu-etal-2025-toolsandbox}
Jiarui Lu, Thomas Holleis, Yizhe Zhang, Bernhard Aumayer, Feng Nan, Haoping
  Bai, Shuang Ma, Shen Ma, Mengyu Li, Guoli Yin, Zirui Wang, and Ruoming Pang.
\newblock {T}ool{S}andbox: A stateful, conversational, interactive evaluation
  benchmark for {LLM} tool use capabilities.
\newblock In Luis Chiruzzo, Alan Ritter, and Lu~Wang, editors, \emph{Findings
  of the Association for Computational Linguistics: NAACL 2025}, pages
  1160--1183, Albuquerque, New Mexico, April 2025. Association for
  Computational Linguistics.
\newblock ISBN 979-8-89176-195-7.
\newblock \doi{10.18653/v1/2025.findings-naacl.65}.
\newblock URL \url{https://aclanthology.org/2025.findings-naacl.65/}.

\bibitem[Naihin et~al.(2023)Naihin, Atkinson, Green, Hamadi, Swift, Schonholtz,
  Kalai, and Bau]{Naihin2023TestingLM}
Silen Naihin, David Atkinson, Marc Green, Merwane Hamadi, Craig Swift, Douglas
  Schonholtz, Adam~Tauman Kalai, and David Bau.
\newblock Testing language model agents safely in the wild.
\newblock \emph{ArXiv}, abs/2311.10538, 2023.
\newblock URL \url{https://api.semanticscholar.org/CorpusID:265281461}.

\bibitem[Nakano et~al.(2021)Nakano, Hilton, Balaji, Wu, Ouyang, Kim, Hesse,
  Jain, Kosaraju, Saunders, et~al.]{nakano2021webgpt}
Reiichiro Nakano, Jacob Hilton, Suchir Balaji, Jeff Wu, Long Ouyang, Christina
  Kim, Christopher Hesse, Shantanu Jain, Vineet Kosaraju, William Saunders,
  et~al.
\newblock Webgpt: Browser-assisted question-answering with human feedback.
\newblock \emph{arXiv preprint arXiv:2112.09332}, 2021.

\bibitem[OpenAI(2025)]{openai2025gptoss120bgptoss20bmodel}
OpenAI.
\newblock gpt-oss-120b \& gpt-oss-20b model card, 2025.
\newblock URL \url{https://arxiv.org/abs/2508.10925}.

\bibitem[Patil et~al.(2025)Patil, Mao, Yan, Ji, Suresh, Stoica, and
  Gonzalez]{patilberkeley}
Shishir~G Patil, Huanzhi Mao, Fanjia Yan, Charlie Cheng-Jie Ji, Vishnu Suresh,
  Ion Stoica, and Joseph~E Gonzalez.
\newblock The berkeley function calling leaderboard (bfcl): From tool use to
  agentic evaluation of large language models.
\newblock In \emph{Forty-second International Conference on Machine Learning},
  2025.

\bibitem[Prabhakar et~al.(2025)Prabhakar, Liu, Zhu, Zhang, Awalgaonkar, Wang,
  Liu, Chen, Hoang, Niebles, et~al.]{prabhakar2025apigen}
Akshara Prabhakar, Zuxin Liu, Ming Zhu, Jianguo Zhang, Tulika Awalgaonkar,
  Shiyu Wang, Zhiwei Liu, Haolin Chen, Thai Hoang, Juan~Carlos Niebles, et~al.
\newblock Apigen-mt: Agentic pipeline for multi-turn data generation via
  simulated agent-human interplay.
\newblock \emph{arXiv preprint arXiv:2504.03601}, 2025.

\bibitem[Qin et~al.(2023)Qin, Liang, Ye, Zhu, Yan, Lu, Lin, Cong, Tang, Qian,
  et~al.]{qin2023toolllm}
Yujia Qin, Shihao Liang, Yining Ye, Kunlun Zhu, Lan Yan, Yaxi Lu, Yankai Lin,
  Xin Cong, Xiangru Tang, Bill Qian, et~al.
\newblock Toolllm: Facilitating large language models to master 16000+
  real-world apis.
\newblock \emph{arXiv preprint arXiv:2307.16789}, 2023.

\bibitem[Reid et~al.(2024)Reid, Savinov, Teplyashin, Lepikhin, Lillicrap,
  Alayrac, Soricut, Lazaridou, Firat, Schrittwieser, et~al.]{reid2024gemini}
Machel Reid, Nikolay Savinov, Denis Teplyashin, Dmitry Lepikhin, Timothy
  Lillicrap, Jean-baptiste Alayrac, Radu Soricut, Angeliki Lazaridou, Orhan
  Firat, Julian Schrittwieser, et~al.
\newblock Gemini 1.5: Unlocking multimodal understanding across millions of
  tokens of context.
\newblock \emph{arXiv preprint arXiv:2403.05530}, 2024.

\bibitem[Ruan et~al.(2024)Ruan, Dong, Wang, Pitis, Zhou, Ba, Dubois, Maddison,
  and Hashimoto]{ruanidentifying}
Yangjun Ruan, Honghua Dong, Andrew Wang, Silviu Pitis, Yongchao Zhou, Jimmy Ba,
  Yann Dubois, Chris~J Maddison, and Tatsunori Hashimoto.
\newblock Identifying the risks of lm agents with an lm-emulated sandbox.
\newblock In \emph{The Twelfth International Conference on Learning
  Representations}, 2024.

\bibitem[Shim et~al.(2025)Shim, Seo, Lim, and Jo]{shim2025tooldial}
Jeonghoon Shim, Gyuhyeon Seo, Cheongsu Lim, and Yohan Jo.
\newblock Tooldial: Multi-turn dialogue generation method for tool-augmented
  language models.
\newblock \emph{arXiv preprint arXiv:2503.00564}, 2025.

\bibitem[Shinn et~al.(2023)Shinn, Cassano, Gopinath, Narasimhan, and
  Yao]{shinn2023reflexion}
Noah Shinn, Federico Cassano, Ashwin Gopinath, Karthik Narasimhan, and Shunyu
  Yao.
\newblock Reflexion: Language agents with verbal reinforcement learning.
\newblock \emph{Advances in Neural Information Processing Systems},
  36:\penalty0 8634--8652, 2023.

\bibitem[Shumailov et~al.(2024)Shumailov, Shumaylov, Zhao, Papernot, Anderson,
  and Gal]{shumailov2024ai}
Ilia Shumailov, Zakhar Shumaylov, Yiren Zhao, Nicolas Papernot, Ross Anderson,
  and Yarin Gal.
\newblock Ai models collapse when trained on recursively generated data.
\newblock \emph{Nature}, 631\penalty0 (8022):\penalty0 755--759, 2024.

\bibitem[Team et~al.(2025)Team, Bai, Bao, Chen, Chen, Chen, Chen, Chen, Chen,
  Chen, et~al.]{team2025kimi}
Kimi Team, Yifan Bai, Yiping Bao, Guanduo Chen, Jiahao Chen, Ningxin Chen,
  Ruijue Chen, Yanru Chen, Yuankun Chen, Yutian Chen, et~al.
\newblock Kimi k2: Open agentic intelligence.
\newblock \emph{arXiv preprint arXiv:2507.20534}, 2025.

\bibitem[Wei et~al.(2022)Wei, Wang, Schuurmans, Bosma, Xia, Chi, Le, Zhou,
  et~al.]{wei2022chain}
Jason Wei, Xuezhi Wang, Dale Schuurmans, Maarten Bosma, Fei Xia, Ed~Chi, Quoc~V
  Le, Denny Zhou, et~al.
\newblock Chain-of-thought prompting elicits reasoning in large language
  models.
\newblock \emph{Advances in neural information processing systems},
  35:\penalty0 24824--24837, 2022.

\bibitem[Wenger(2024)]{wenger2024ai}
Emily Wenger.
\newblock Ai produces gibberish when trained on too much ai-generated data,
  2024.

\bibitem[Xie et~al.(2024)Xie, Zhang, Chen, Li, Zhao, Cao, Hua, Cheng, Shin,
  Lei, et~al.]{xie2024osworld}
Tianbao Xie, Danyang Zhang, Jixuan Chen, Xiaochuan Li, Siheng Zhao, Ruisheng
  Cao, Toh~J Hua, Zhoujun Cheng, Dongchan Shin, Fangyu Lei, et~al.
\newblock Osworld: Benchmarking multimodal agents for open-ended tasks in real
  computer environments.
\newblock \emph{Advances in Neural Information Processing Systems},
  37:\penalty0 52040--52094, 2024.

\bibitem[Xu et~al.(2025)Xu, Soria, Tan, Roy, Agrawal, Poovendran, and
  Panda]{xu2025toucan}
Zhangchen Xu, Adriana~Meza Soria, Shawn Tan, Anurag Roy, Ashish~Sunil Agrawal,
  Radha Poovendran, and Rameswar Panda.
\newblock Toucan: Synthesizing 1.5 m tool-agentic data from real-world mcp
  environments.
\newblock \emph{arXiv preprint arXiv:2510.01179}, 2025.

\bibitem[Yang et~al.(2025{\natexlab{a}})Yang, Li, Yang, Zhang, Hui, Zheng, Yu,
  Gao, Huang, Lv, et~al.]{yang2025qwen3}
An~Yang, Anfeng Li, Baosong Yang, Beichen Zhang, Binyuan Hui, Bo~Zheng, Bowen
  Yu, Chang Gao, Chengen Huang, Chenxu Lv, et~al.
\newblock Qwen3 technical report.
\newblock \emph{arXiv preprint arXiv:2505.09388}, 2025{\natexlab{a}}.

\bibitem[Yang et~al.(2025{\natexlab{b}})Yang, Le, Xing, An, Chen, Zhao, Song,
  and Zhang]{yang2025toolmindtechnicalreportlargescale}
Chen Yang, Ran Le, Yun Xing, Zhenwei An, Zongchao Chen, Wayne~Xin Zhao, Yang
  Song, and Tao Zhang.
\newblock Toolmind technical report: A large-scale, reasoning-enhanced tool-use
  dataset, 2025{\natexlab{b}}.
\newblock URL \url{https://arxiv.org/abs/2511.15718}.

\bibitem[Yang et~al.(2024)Yang, Yang, Zhang, Hui, Zheng, Yu, Li, Liu, Huang,
  Dong, Wei, Lin, Yang, Tu, Zhang, Yang, Yang, Zhou, Lin, Dang, Lu, Bao, Yang,
  Yu, Li, Xue, Zhang, Zhu, Men, Lin, Li, Xia, Ren, Ren, Fan, Su, Zhang, Wan,
  Liu, Cui, Zhang, Qiu, Quan, and Wang]{Yang2024Qwen25TR}
Qwen~An Yang, Baosong Yang, Beichen Zhang, Binyuan Hui, Bo~Zheng, Bowen Yu,
  Chengyuan Li, Dayiheng Liu, Fei Huang, Guanting Dong, Haoran Wei, Huan Lin,
  Jian Yang, Jianhong Tu, Jianwei Zhang, Jianxin Yang, Jiaxin Yang, Jingren
  Zhou, Junyang Lin, Kai Dang, Keming Lu, Keqin Bao, Kexin Yang, Le~Yu, Mei Li,
  Mingfeng Xue, Pei Zhang, Qin Zhu, Rui Men, Runji Lin, Tianhao Li, Tingyu Xia,
  Xingzhang Ren, Xuancheng Ren, Yang Fan, Yang Su, Yi-Chao Zhang, Yunyang Wan,
  Yuqi Liu, Zeyu Cui, Zhenru Zhang, Zihan Qiu, Shanghaoran Quan, and Zekun
  Wang.
\newblock Qwen2.5 technical report.
\newblock \emph{ArXiv}, abs/2412.15115, 2024.
\newblock URL \url{https://api.semanticscholar.org/CorpusID:274859421}.

\bibitem[Yao et~al.(2023)Yao, Zhao, Yu, Du, Shafran, Narasimhan, and
  Cao]{yao2023react}
Shunyu Yao, Jeffrey Zhao, Dian Yu, Nan Du, Izhak Shafran, Karthik Narasimhan,
  and Yuan Cao.
\newblock React: Synergizing reasoning and acting in language models.
\newblock In \emph{International Conference on Learning Representations
  (ICLR)}, 2023.

\bibitem[Yao et~al.(2024)Yao, Shinn, Razavi, and Narasimhan]{yao2024tau}
Shunyu Yao, Noah Shinn, Pedram Razavi, and Karthik Narasimhan.
\newblock {$\tau$}-bench: A benchmark for tool-agent-user interaction in
  real-world domains.
\newblock \emph{arXiv preprint arXiv:2406.12045}, 2024.

\bibitem[Yin et~al.(2025)Yin, Wang, Hsu, Yan, Jiang, Chen, Gu, Le, Chang, Lee,
  et~al.]{yin2025magnet}
Fan Yin, Zifeng Wang, I~Hsu, Jun Yan, Ke~Jiang, Yanfei Chen, Jindong Gu, Long~T
  Le, Kai-Wei Chang, Chen-Yu Lee, et~al.
\newblock Magnet: Multi-turn tool-use data synthesis and distillation via graph
  translation.
\newblock \emph{arXiv preprint arXiv:2503.07826}, 2025.

\bibitem[Zeng et~al.(2025{\natexlab{a}})Zeng, Lv, Zheng, Hou, Chen, Xie, Wang,
  Yin, Zeng, Zhang, et~al.]{zeng2025glm}
Aohan Zeng, Xin Lv, Qinkai Zheng, Zhenyu Hou, Bin Chen, Chengxing Xie, Cunxiang
  Wang, Da~Yin, Hao Zeng, Jiajie Zhang, et~al.
\newblock Glm-4.5: Agentic, reasoning, and coding (arc) foundation models.
\newblock \emph{arXiv preprint arXiv:2508.06471}, 2025{\natexlab{a}}.

\bibitem[Zeng et~al.(2025{\natexlab{b}})Zeng, Liu, Wang, Li, Mi, Wang, Shang,
  Jiang, and Liu]{zeng2025toolace}
Xingshan Zeng, Weiwen Liu, Lingzhi Wang, Liangyou Li, Fei Mi, Yasheng Wang,
  Lifeng Shang, Xin Jiang, and Qun Liu.
\newblock Toolace-mt: Non-autoregressive generation for agentic multi-turn
  interaction.
\newblock \emph{arXiv preprint arXiv:2508.12685}, 2025{\natexlab{b}}.

\bibitem[Zhang et~al.(2025{\natexlab{a}})Zhang, Hoang, Zhu, Liu, Wang,
  Awalgaonkar, Prabhakar, Chen, Yao, Liu, Tan, Niebles, Heinecke, Wang,
  Savarese, and Xiong]{zhang-etal-2025-actionstudio}
Jianguo Zhang, Thai Hoang, Ming Zhu, Zuxin Liu, Shiyu Wang, Tulika Awalgaonkar,
  Akshara Prabhakar, Haolin Chen, Weiran Yao, Zhiwei Liu, Juntao Tan,
  Juan~Carlos Niebles, Shelby Heinecke, Huan Wang, Silvio Savarese, and Caiming
  Xiong.
\newblock {A}ction{S}tudio: A lightweight framework for data and training of
  large action models.
\newblock In Christos Christodoulopoulos, Tanmoy Chakraborty, Carolyn Rose, and
  Violet Peng, editors, \emph{Proceedings of the 2025 Conference on Empirical
  Methods in Natural Language Processing}, pages 21488--21502, Suzhou, China,
  November 2025{\natexlab{a}}. Association for Computational Linguistics.
\newblock ISBN 979-8-89176-332-6.
\newblock \doi{10.18653/v1/2025.emnlp-main.1090}.
\newblock URL \url{https://aclanthology.org/2025.emnlp-main.1090/}.

\bibitem[Zhang et~al.(2025{\natexlab{b}})Zhang, Lan, Zhu, Liu, Hoang, Kokane,
  Yao, Tan, Liu, Feng, Niebles, Heinecke, Wang, Savarese, and
  Xiong]{zhang-etal-2025-xlam}
Jianguo Zhang, Tian Lan, Ming Zhu, Zuxin Liu, Thai Hoang, Shirley Kokane,
  Weiran Yao, Juntao Tan, Zhiwei Liu, Yihao Feng, Juan~Carlos Niebles, Shelby
  Heinecke, Huan Wang, Silvio Savarese, and Caiming Xiong.
\newblock x{LAM}: A family of large action models to empower {AI} agent
  systems.
\newblock In Luis Chiruzzo, Alan Ritter, and Lu~Wang, editors,
  \emph{Proceedings of the 2025 Conference of the Nations of the Americas
  Chapter of the Association for Computational Linguistics: Human Language
  Technologies (Volume 1: Long Papers)}, pages 11583--11597, Albuquerque, New
  Mexico, April 2025{\natexlab{b}}. Association for Computational Linguistics.
\newblock ISBN 979-8-89176-189-6.
\newblock \doi{10.18653/v1/2025.naacl-long.578}.
\newblock URL \url{https://aclanthology.org/2025.naacl-long.578/}.

\bibitem[Zhang et~al.(2023)Zhang, Shen, Lu, and Zhuang]{zhang2023data}
Wenqi Zhang, Yongliang Shen, Weiming Lu, and Yueting Zhuang.
\newblock Data-copilot: Bridging billions of data and humans with autonomous
  workflow.
\newblock \emph{arXiv preprint arXiv:2306.07209}, 2023.

\bibitem[Zheng et~al.(2024)Zheng, Zhang, Zhang, Ye, Luo, Feng, and
  Ma]{zheng2024llamafactory}
Yaowei Zheng, Richong Zhang, Junhao Zhang, Yanhan Ye, Zheyan Luo, Zhangchi
  Feng, and Yongqiang Ma.
\newblock Llamafactory: Unified efficient fine-tuning of 100+ language models.
\newblock In \emph{Proceedings of the 62nd Annual Meeting of the Association
  for Computational Linguistics (Volume 3: System Demonstrations)}, Bangkok,
  Thailand, 2024. Association for Computational Linguistics.
\newblock URL \url{http://arxiv.org/abs/2403.13372}.

\bibitem[Zhou et~al.(2025)Zhou, Wang, Lu, Shi, Tie, Xu, Chen, Zhou, Gong, and
  Sun]{Zhou2025SafeAgentSL}
Xueyang Zhou, Weidong Wang, Lin Lu, Jiawen Shi, Guiyao Tie, Yongtian Xu, Lixing
  Chen, Pan Zhou, Neil~Zhenqiang Gong, and Lichao Sun.
\newblock Safeagent: Safeguarding llm agents via an automated risk simulator.
\newblock 2025.
\newblock URL \url{https://api.semanticscholar.org/CorpusID:278885968}.

\end{thebibliography}
